\documentclass[sigconf]{acmart}
\AtBeginDocument{%
  }

\copyrightyear{2026}
\acmYear{2026}
\setcopyright{cc}
\setcctype{by}
\acmConference[WWW '26]{Proceedings of the ACM Web Conference 2026}{April 13--17, 2026}{Dubai, United Arab Emirates}
\acmBooktitle{Proceedings of the ACM Web Conference 2026 (WWW '26), April 13--17, 2026, Dubai, United Arab Emirates}
\acmPrice{}
\acmDOI{10.1145/3774904.3792145}
\acmISBN{979-8-4007-2307-0/2026/04}

\makeatletter
\gdef\@copyrightpermission{
  \begin{minipage}{0.3\columnwidth}
   \href{https://creativecommons.org/licenses/by/4.0/}{\includegraphics[width=0.90\textwidth]{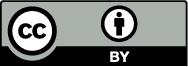}}
  \end{minipage}\hfill
  \begin{minipage}{0.7\columnwidth}
   \href{https://creativecommons.org/licenses/by/4.0/}{This work is licensed under a Creative Commons Attribution International 4.0 License.}
  \end{minipage}
  \vspace{0pt}
}
\makeatother

\usepackage{microtype}
\usepackage{hyperref}
\usepackage{url}
\usepackage{booktabs}
\usepackage{adjustbox}
\usepackage{multirow}
\usepackage{multicol}
\usepackage{xspace}
\usepackage{listings}
\usepackage{enumitem}
\usepackage{algorithm}
\usepackage{algpseudocode}

\begin{document}

\title{Pathways of Thoughts: Multi-Directional Thinking for Long-form Personalized Question Answering}


\author{Alireza Salemi}
\affiliation{%
  \institution{University of Massachusetts Amherst}
  \city{Amherst}
  \state{MA}
  \country{USA}}
\email{asalemi@cs.umass.edu}

\author{Cheng Li}
\affiliation{%
  \institution{Google DeepMind}
  \city{Mountain View}
  \state{CA}
  \country{USA}}
\email{chgli@google.com}

\author{Mingyang Zhang}
\affiliation{%
  \institution{Google DeepMind}
  \city{Mountain View}
  \state{CA}
  \country{USA}}
\email{mingyang@google.com}

\author{Qiaozhu Mei}
\affiliation{%
  \institution{University of Michigan}
  \city{Ann Arbor}
  \state{MI}
  \country{USA}}
\email{qmei@umich.edu}

\author{Zhuowan Li}
\affiliation{%
  \institution{Google DeepMind}
  \city{Mountain View}
  \state{CA}
  \country{USA}}
\email{zhuowan@google.com}

\author{Spurthi Amba Hombaiah}
\affiliation{%
  \institution{Google DeepMind}
  \city{Mountain View}
  \state{CA}
  \country{USA}}
\email{spurthiah@google.com}

\author{Weize Kong}
\affiliation{%
  \institution{Google DeepMind}
  \city{Mountain View}
  \state{CA}
  \country{USA}}
\email{weize@google.com}

\author{Tao Chen}
\affiliation{%
  \institution{Google DeepMind}
  \city{Mountain View}
  \state{CA}
  \country{USA}}
\email{taochen@google.com}

\author{Hamed Zamani}
\affiliation{%
  \institution{University of Massachusetts Amherst}
  \city{Amherst}
  \state{MA}
  \country{USA}}
\email{zamani@cs.umass.edu}

\author{Michael Bendersky}
\affiliation{%
  \institution{Google DeepMind}
  \city{Mountain View}
  \state{CA}
  \country{USA}}
\email{bemike@google.com}
\settopmatter{authorsperrow=4}


\newcommand{\rmdpwithsearch}{Pathways of Thoughts\xspace}
\newcommand{\rmdpwithsearchshort}{PoT\xspace}
\newcommand{\dataset}{LaMP-QA\xspace}
\newcommand{\datasetshort}{LaMP-QA\xspace}
\renewcommand{\algorithmicrequire}{\textbf{Input:}}
\renewcommand{\algorithmicensure}{\textbf{Output:}}
\newcommand{\cheng}[1]{\textcolor{blue}{[Cheng: #1]}}
\newcommand{\weize}[1]{\textcolor{purple}{[Weize: #1]}}
\newcommand{\hamed}[1]{\textcolor{orange}{[Hamed: #1]}}
\newcommand{\alireza}[1]{\textcolor{olive}{[Alireza: #1]}}

\begin{abstract}
Personalization is well studied in search and recommendation, but personalized question answering remains underexplored due to challenges in inferring preferences from long, noisy, implicit contexts and generating responses that are both accurate and aligned with user expectations. To address this, we propose \textit{\rmdpwithsearch (\rmdpwithsearchshort)}, an inference-stage method that applies to any large language model (LLM) without task-specific fine-tuning. \rmdpwithsearchshort models the thinking as an iterative decision process, where the model dynamically selects among cognitive operations such as reasoning, revision, personalization, and clarification. This enables exploration of multiple reasoning trajectories, producing diverse candidate responses that capture different perspectives. {\rmdpwithsearchshort} then aggregates and reweights these candidates according to inferred user preferences, yielding a final personalized response that benefits from the complementary strengths of diverse reasoning paths. Experiments on the LaMP-QA benchmark show that {\rmdpwithsearchshort} consistently outperforms competitive baselines, achieving up to a 10.8\% relative improvement. Human evaluation further validates these improvements, with annotators preferring {\rmdpwithsearchshort} in 66\% of cases compared to the best-performing baseline and reporting ties in 15\% of cases.
\vspace{-0.3cm}
\end{abstract}



\settopmatter{printacmref=true}
\maketitle

\section{Introduction}
\label{sec:intro}

Personalizing large language models (LLMs) has become increasingly important due to applications in content generation, writing assistance, education, and recommendation \citep{lamp, longlamp, mysore2023pearl, li2023teach, naumov2019deep}. By conditioning outputs on user preferences, context, and needs, personalization improves engagement, satisfaction, and overall effectiveness \citep{salemi2025reasoningenhancedselftraininglongformpersonalized}. It is particularly valuable in information-seeking systems such as search engines and question answering (QA) \citep{lamp-qa, salemi2025learningnaturallanguagefeedback}, where appropriate responses depend on user-specific factors like background knowledge and learning style. Despite its clear relevance to web search and emerging systems (e.g., AI-powered search modes), personalization in QA remains relatively underexplored.

\begin{figure*}
    \centering
    \includegraphics[width=0.85\textwidth]{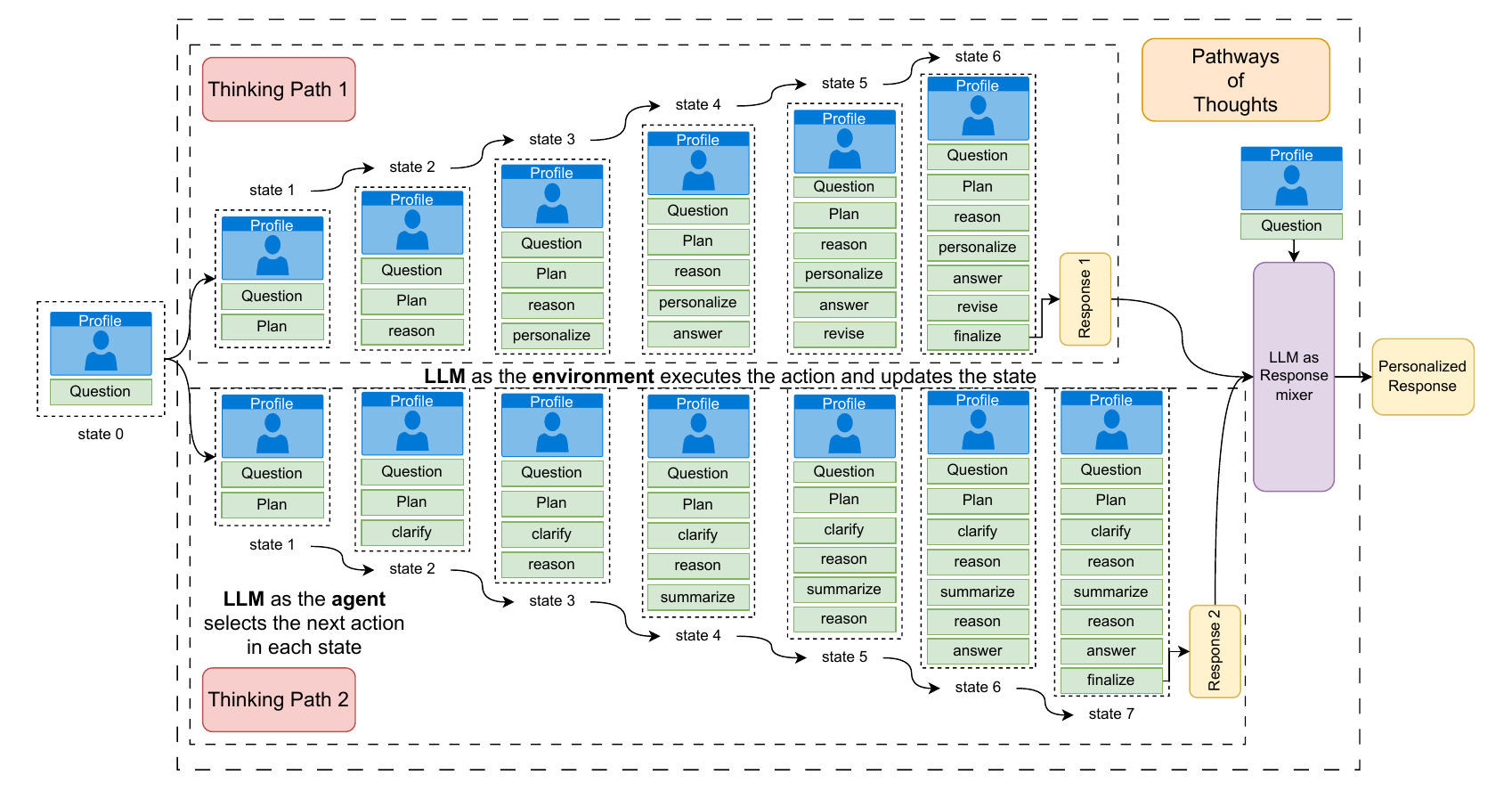}
    \vspace{-0.5cm}
    \caption{\rmdpwithsearch overview. \rmdpwithsearchshort explores multiple directions of thinking about answering the question, generates a response for each, and mix them into a final response.}
    \label{fig:overview}
    \vspace{-0.5cm}
\end{figure*}

Augmenting LLM inputs with the user's personalized context have proven effective for personalized generation \citep{lamp, rspg, longlamp, li2023teach}. This can be further enhanced by training the LLM to perform reasoning over the personalized context to extract the user's preferences before response generation \citep{salemi2025reasoningenhancedselftraininglongformpersonalized}. However, in most real-world cases, fine-tuning the LLM is often infeasible for two key reasons. On one hand, many LLMs are black boxes with frozen parameters, offering no control over their internal mechanisms. On the other hand, even with full access to parameters, the fine-tuning process requires 
substantial amount of personal data and computational resources for individual users. Consequently, a practical way to influence the model's output is by applying advanced strategies during inference. Recognizing this, our paper focuses on this scenario and proposes an \textbf{inference-stage strategy} to achieve better personalization.

Developing effective personalization models presents its own unique challenges. First, the quality of a response in personalized QA is determined by the user who asks the question, which is influenced by their preferences, context, and background. Therefore, even when two answers are factually accurate, one may be preferred by the user. Second, user preferences are often not explicitly stated and must be inferred from personalized context, such as profiles or prior interactions. This context is frequently lengthy and noisy, making it difficult for a model to extract relevant signals.
These issues highlight that simply augmenting an LLM's input with user context is often insufficient. Effective personalization requires the model to perform reasoning over noisy, implicit user data to deduce user preferences before generating a response.


To address the aforementioned challenges of personalized QA, this paper introduces \textit{\rmdpwithsearch (\rmdpwithsearchshort)}, demonstrated in Figure~\ref{fig:overview}. This approach, which is generic and can be applied to any LLM without additional training, enables the model to explore multiple thinking pathways when generating responses, each leading to a distinct response.
We formalize each thinking pathway as a Markov Decision Process (MDP), where the LLM serves as both the agent and the environment. At each step, the LLM, acting as the agent, selects the most suitable action from a predefined set of cognitive operations, such as planning, reasoning, personalizing, and revising. The LLM then assumes the role of the environment, executing the chosen action and updating the state with newly acquired knowledge. This iterative process allows the model to refine its understanding of the user's intent, extract key preferences from the user profile, and explore diverse strategies for answering the question. The process continues until the model determines that no further actions are needed, producing an initial set of responses. To generate the final personalized response, the LLM mixes outputs from different thinking pathways into a single response, aligning them with the user's preferences extracted from their profile. \rmdpwithsearch addresses the challenges of LLMs in personalized question answering by allowing the model to iteratively think about the next step in responding to the question and how the lengthy user context can be effectively utilized. Additionally, it explores different diverse directions and paths, considering a range of possible solutions for answering the question. Finally, generating a response by aggregating all pathways' responses ensures that the output captures the strengths of all explored pathways, resulting in a more personalized and aligned response. 
We conduct our experiments on the  \textit{\datasetshort} benchmark \cite{lamp-qa}--a recent large-scale benchmark designed for evaluating personalized QA in three diverse domains: (1) Art \& Entertainment, (2) Lifestyle \& Personal Development, and (3) Society \& Culture. Our results demonstrate that {\rmdpwithsearchshort} outperforms competitive baselines across all categories, with up to $10.8\%$ statistically significant improvement on average compared to the best-performing baselines. Additionally, we investigate the impact of pathway length, the number of pathways, and various methods for diversifying pathways and aggregating responses on performance. We show that {\rmdpwithsearchshort} generalizes to different long-context backbone LLMs without requiring additional training. Finally, human evaluation of comparing outputs side by side shows that \textit{\rmdpwithsearchshort} better satisfies users' information needs in $66\%$ of cases, compared to $19\%$ for the best-performing baseline.

\section{Related Work}

\subsubsection*{\textbf{LLM Personalization}}

Personalization plays a central role in search, recommendation, and text generation applications \citep{10.1145/2702123.2702503, 10.1145/1462198.1462203, naumov2019deep, lamp}. To enable personalization in LLMs, \citet{lamp} introduced a retrieval-augmented generation framework alongside the LaMP benchmark for short-form text, which was later extended to long-form content generation with LongLaMP \citep{longlamp} and more recently to personalized QA with LaMP-QA \citep{lamp-qa}. Personalized assistants have also been studied in several recent works \citep{li2023teach, mysore2023pearl, lu2024corporate, zhang-etal-2024-llm-based}. Proposed approaches span a variety of techniques, including training retrieval models with user feedback \citep{rspg}, optimizing LLMs with personalized supervision \citep{jang2023personalized}, and generating personalized prompts \citep{Li_2024}. Parameter-efficient fine-tuning has emerged as another direction \citep{tan2024personalized}, including its integration with retrieval-augmented methods \citep{peft-rag-personalization}. Beyond model adaptation, reasoning and self-training strategies have been shown to improve personalization in text generation \citep{salemi2025reasoningenhancedselftraininglongformpersonalized}. More recently, learning from natural language feedback has proven effective for adapting QA systems to individual users \citep{salemi2025learningnaturallanguagefeedback}. Nevertheless, strategies for improving personalized QA at test time---without incurring the cost of additional training---remain underexplored, which is the focus of this work.

\subsubsection*{\textbf{Reasoners \& Agents for Problem Solving}}
Reasoning enables models to solve problems with step-by-step processing, known as chain-of-thought (CoT), enhancing performance in complex tasks like mathematical problem-solving, logical reasoning, and commonsense understanding \citep{cot, liu2023logicot, 10.1007/978-981-97-7232-2_13}. Training LLMs to perform intermediate reasoning steps before generating responses has been effective, as seen in GPT-o1 \citep{openai_o1} and DeepSeek-R1 \citep{deepseek-r1}. Additionally, agentic and multi-turn or multi-agent systems that interact with external tools \cite{search-r1}, code \citep{liu2025toolplanner, xie2024a, fu2024autoguide}, simulated environments \citep{hao2023reasoning, wang2024qimprovingmultistepreasoning, park2024ensemblinglargelanguagemodels, yao2022react}, or with other agents \cite{multi-agent}  have been explored for complex problem-solving. Markov Decision Processes (MDP) have shown promise in reasoning for games and mathematical problems \citep{hao2023reasoning, wang2024qimprovingmultistepreasoning}. However, these methods have not been applied to free-form text generation, particularly personalization which requires user-specific responses rather than objectively correct answers. This study extends these methods to personalized text generation by integrating them with graph-based search in place of traditional tree search in MDPs, enabling the model to produce responses that reflect the aggregated quality of all explored candidate solutions rather than relying on a single path.

\subsubsection*{\textbf{Scaling Inference Compute}}
Recent advances in LLM reasoning for logic and mathematics show that increasing compute resources during inference can significantly enhance performance \citep{snell2024scalingllmtesttimecompute, chen2024simpleprovablescalinglaw}. This allows LLMs to explore answer spaces more thoroughly, improving accuracy in logical reasoning, math problem-solving, and code generation \citep{bi2024forestofthoughtscalingtesttimecompute, zhang2024scalingllminferenceoptimized, brown2024largelanguagemonkeysscaling}. While previous research has focused on math, coding, and logic, its potential for free-form text generation, especially personalized question answering, is underexplored. Our work addresses this by using increased inference compute to better navigate response spaces, enhancing personalization.

\section{Problem Formulation}
\label{sec:problem-formulation}

We assume a question \( x_u \) originates from a user \( u \), who has a set of \( n_u \) personalized information elements, denoted as \( P_u = \{p_i\}_{i=1}^{n_u} \). Following previous work that utilizes long-term user history as the user profile \citep{lamp-qa, lamp, se-pqa}, \( P_u \) consists of the user's previously asked questions along with the their question narrative. The objective is to use this personalized information, along with the question \( x_u \), to generate a personalized response using an LLM \( \pi \), formulated as \( \hat{y}_u = \pi(x_u, P_u) \). To evaluate the personalized response generated for the user \( u \)'s question, following \citet{lamp-qa}, we assume access to a set of aspects \( E_{x_u} = \{e_i\}_{i=1}^{|E_{x_u}|} \) that user \( u \) expects to be addressed in the response, extracted from a question narrative $r_{x_u}$ written by the user. These aspects are used solely for evaluation and are not accessible to the model during response generation. Finally, a metric \( \mu(x_u, \hat{y}_u, E_{x_u}, r_u) \) is used to score the generated response based on the extent to which these aspects are covered.

\section{\rmdpwithsearch}
\label{sec:method}

Personalization is inherently user-specific, meaning that a response generation strategy effective for one user may not perform well for another. Consequently, relying on a single solution to generate a response for a user may not always yield the most preferred outcome for each individual user. To address this limitation, as well as those discussed in Section~\ref{sec:intro}, we propose \textit{\rmdpwithsearch (\rmdpwithsearchshort)}, illustrated in Figure~\ref{fig:overview}. In this framework, the LLM explores multiple thinking pathways, each corresponding to a distinct line of thought that leads to a potential response. We model each thinking pathway as an MDP, where the LLM serves as both the agent and the environment. At each step, the LLM (acting as the agent) selects an action from a set of fundamental cognitive operations such as planning, reasoning, clarification, or revising. The same LLM (now acting as the environment) then executes the chosen action, updating the current state with newly acquired information about the question and user. This iterative process continues until convergence, producing a final response for each pathway. To generate a single personalized response, the LLM aggregates the responses obtained from different pathways, integrating their complementary strengths while aligning with the user’s preferences inferred from their profile. The complete procedure is detailed in Algorithm~\ref{alg:pot}, and the next section elaborates on this method.

\begin{algorithm}
\caption{\rmdpwithsearch Implementation using MDP and Global Search.}\label{alg:pot}
\begin{algorithmic}[1]
\Require prompt $x_u$, profile $P_u$, max path length $T$, number of pathways $N$, backbone LLM $\pi$
\Ensure personalized output ${\hat{y}_{x_u}}$
\State $R_{\text{PoT}} = \{\}$ \Comment{Set of all responses from different pathways}
\For{$i = 0$ to $N$} \Comment{Generate $N$ pathways of thoughts}
    \State $t = 0$
    \State $s_t = \text{init\_prompt}(x_u, P_u)$ \Comment{Prompt in Figure~\ref{fig:init-prompt}}
    \While{$t < T$} \Comment{Thinking for maximum $T$ steps}
    \State $A_{s_t} = \text{possible\_actions}(s_t, t)$ \Comment{Find the possible actions for state $s_t$.}
    \State $\bar{a}_{p} = \text{action\_selection\_prompt}(s_t, A_{s_t})$ \Comment{Prompt in Figure~\ref{fig:action-selection-prompt}}
    \State $a_t = \pi(s_t;\bar{a}_{p})$ \Comment{Selecting the action (LLM as agent)}
    \State $\hat{y}_t = \pi(s_t;\bar{a}_{p};a_t)$ \Comment{Performing the action (LLM as environment)}
    \State $s_{t+1} = s_t;\bar{a}_{p};a_t;\hat{y}_t$ \Comment{new state by concatenating input/outputs in this time step}
    \If{$a_t$ = finalize}
        \State $R_{\text{PoT}} = R_{\text{PoT}} \cup \{\hat{y}_t\}$ \Comment{Capturing the final response of this pathway}
        \State \textbf{break}
    \EndIf
    \State $t = t + 1$ \Comment{updating the step variable}
    \EndWhile
\EndFor
\State $\bar{I}_{p} = \text{important\_aspects\_prompt}(P_u)$ \Comment{Prompt in Figure~\ref{fig:important-aspects-user-prompt}}
\State $I_u = \pi(\bar{I}_{p})$ \Comment{Extracting important aspects from user profile}
\State $\bar{y}_p = \text{mixture\_prompt}(I_u, x_u, R_{\text{PoT}})$ \Comment{Prompt in Figure~\ref{fig:mixture-of-N-prompt}}
\State ${\hat{y}_{x_u}} = \pi(\bar{y}_p)$ \Comment{Combining responses based on important aspects}
\State\Return ${\hat{y}_{x_u}}$ \Comment{Returning final response}
\end{algorithmic}
\end{algorithm}

\begin{figure*}
    \centering
    \includegraphics[width=0.85\textwidth]{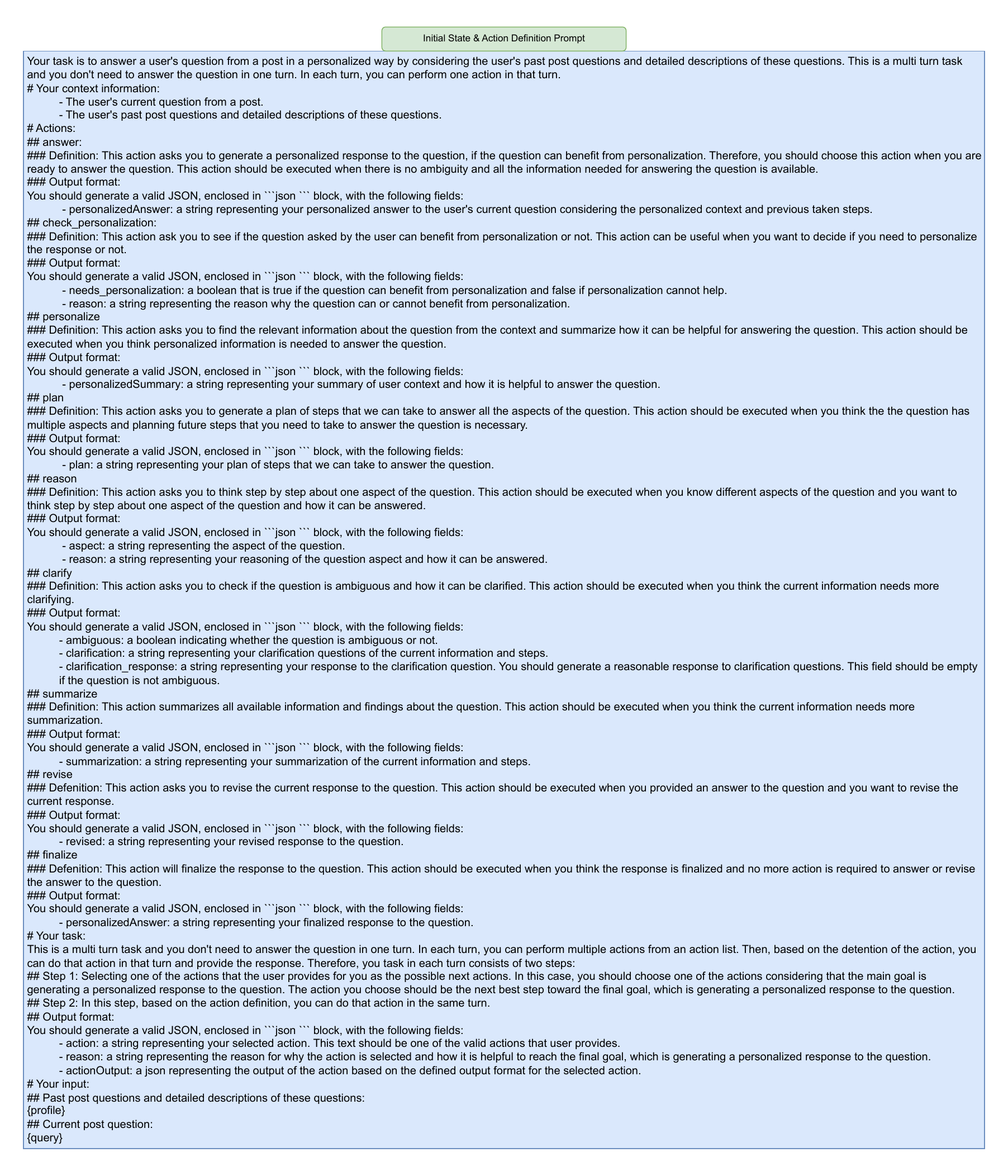}
    \vspace{-0.8cm}
    \caption{Initial state and actions definition prompt used in \rmdpwithsearch.}
    \label{fig:init-prompt}
    \vspace{-0.4cm}
\end{figure*}

\subsection{Thinking as a Markov Decision Process}

The thinking process in the human brain involves using and integrating a set of fundamental cognitive skills---e.g., reasoning, planning, and revision---that collectively form the foundation of structured thought \citep{thinking-process-1, thinking-process-2}. Inspired by theories about sequential decision making in the human brain, we model the thinking process of LLMs as an MDP. Given the current thoughts and steps taken to generate a response, the model selects the next fundamental action from a predefined set to enhance its representation of the question, refine its reasoning, and improve the generated response.

An MDP is defined as a quadruple \( (S, A_{s_t}, P_{a_t}, R_{a_t}) \), where \( S \) is the set of states, \( A_{s_t} \) represents the set of available actions in state \( s_t \), \( P_{a_t}(s_{t+1} \mid s_t) \) defines the probability of transitioning from state \( s_t \) to \( s_{t+1} \) given action \( {a_t} \), and \( R_{a_t}(s_{t+1}, s_t) \) specifies the reward function, which assigns a reward for taking action \( {a_t} \) in state \( s_t \) and transitioning to \( s_{t+1} \). In this setting, the agent \( \pi \) aims to assign a probability \( p_{\pi}({a_t} | s_t) \) to each action based on the current state \( s_t \). 

To generate a personalized response for the question \( x_u \) from the user \( u \), we assume the agent \( \pi \) starts from an initial state \( s_0 \), which consists of the question \( q_u \) and the user profile \( P_u \). The agent iteratively selects the most probable action at each step using the prompt shown in Figure~\ref{fig:action-selection-prompt} in Appendix \ref{app:imp-details}, continuing this process until it either chooses a halting action (as determined by the LLM) or reaches the maximum number of allowed actions, denoted as \( T \). Since our task involves text generation and lacks an external environment enforcing termination, the agent itself determines when to stop or continues until the predefined action limit is reached. See =
Algorithm~\ref{alg:pot} (lines 3-17) for details. The following subsections outlines the definitions of each MDP component in our framework:

\subsubsection*{\textbf{Set of States ($S$):}}

The state is defined as the concatenation of the inputs, outputs, and decisions of the model into a sequence of tokens using a predefined prompt format. This representation captures the full context of the system like a conversation. The initial state of the system consists of the user’s question \( x_u \), the user profile \( P_u \), and a set of instructions specifying the actions and the model’s expected behavior for each action, as shown in Figure~\ref{fig:init-prompt}. Given this setup, the set of all possible states corresponds to the set of all textual sequences that this initial prompt is their prefix. As a result, the state space forms an uncountable discrete set.

\subsubsection*{\textbf{Set of Possible Actions for State $s_t$ ($A_{s_t}$):}}

We consider the following fundamental actions that the agent can choose from in thinking process: \textit{answering}, \textit{planning}, \textit{personalization}, \textit{personalizing}, \textit{reasoning}, \textit{clarifying}, \textit{summarizing}, \textit{revising}, and \textit{finalizing}.\footnote{This action, \textit{finalizing}, also referred to as halting, terminates the process.} The detailed definition of each action and output format of them is provided in Appendix~\ref{app:imp-details-action-def}. In the initial state, the agent is restricted to selecting the \textit{planning} action, as it must first plan the steps required to address the user's question. In subsequent states, the agent can choose from all the actions; however, the model can only select the \textit{revising} action if it has previously selected the \textit{answering} action, as revision typically follows the generation of a response. 

\subsubsection*{\textbf{Transitioning from State \( s_t \) to \( s_{t+1} \) with Action \( {a_t} \) ($P_{a_t}(s_{t+1} \mid s_t)$):}}

We use the LLM as both the agent and the environment simultaneously: when it selects an action, it acts as the agent, and when it executes the action, it acts as the environment. This dual role allows the LLM to control the flow of thinking and performing it simultaneously. Accordingly, with the agent decision to perform action \( a_t \) at time step \( t \), the LLM receives the state \( s_t \) and action \( a_t \) as input and generates the output \( \hat{y}_t \). The concatenation of \( s_t \), \( a_t \), and \( \hat{y}_t \) forms the next state \( s_{t+1} \). Thus, the transition probability \( P_{a_t}(s_{t+1} \mid s_t) \) is defined by the LLM itself.

\subsubsection*{\textbf{Reward Function ($R_a(s_{t+1}, s_t)$)}}
Since our agent is used without training, an explicit reward function is not required; we assume the agent makes reasonable choices. However, in general, an outcome reward function can be employed to assign rewards to actions based on the evaluation of the LLM's final response to the question.

\subsection{Multi-Directional Thinking by Diverse Planning and Mixture-of-Responses}
\label{sec:multi-directional}

\begin{table*}
    \centering
    \caption{Performance on \datasetshort. $\dagger$ and $\ddagger$ show statistically significant improvement over the single and multiple inference best baseline, respectively, using t-test ($p < 0.05$).  Gemini 1.5 Pro has been used as the backbone LLM for all the methods.}
    \vspace{-0.4cm}
    \adjustbox{max width=0.82\textwidth}{
    \begin{tabular}{ll|ccc|c}
        \toprule
        & \multirow{2}{*}{\textbf{Method}} & \textbf{Arts \&} & \textbf{Lifestyle \& Personal} & \textbf{Society \&} & \textbf{Average} \\
        & & \textbf{Entertainment} & \textbf{Development} & \textbf{Culture} & \textbf{(macro)} \\
        \midrule
        \multicolumn{6}{c}{Single Inference} \\
        \midrule
        (1) & No Personalization & 0.1741	& 0.2863 & 0.2996 & 0.2533 \\
        (2) & In-Context Personalization & 0.1722	& 0.3022 & 0.3482 & 0.2742 \\
        (3) & In-Context Personalization w/ CoT & 0.2119 & 0.3862 & 0.4230 & 0.3403 \\
        \midrule
        \midrule
        (4) & \textbf{\rmdpwithsearch ($N=1$, $T=8$)} & \textbf{0.2499$^{\dagger}$} & \textbf{0.3966} & \textbf{0.4490$^{\dagger}$} & \textbf{0.3651$^{\dagger}$}	\\
        \midrule
        \multicolumn{6}{c}{Multiple Inference} \\
        \midrule
        {(5)} & In-Context Personalization w/ Best-of-32 & {0.2248} & {0.3753} &	{0.4303} & {0.3434} \\
    
        {(6)} & In-Context Personalization w/ CoT \& Best-of-32 & {0.2607} & {0.4380} & {0.4789} & {0.3925} \\
        
        {(7)} & In-Context Personalization w/ Mix-of-32 & 0.2310 & 0.3781 & 0.4411 & 0.3500 \\
        
        {(8)} & In-Context Personalization w/ CoT \& Mix-of-32 & 0.2689 & 0.4443 & 0.4891 & 0.4007 \\
        
        {(9)} & Tree of Thoughts ($K=32$, $B=2$) & {0.2054} & {0.4178} & {0.4546} & {0.3592} \\
        \midrule
        \midrule
        (10) & \textbf{\rmdpwithsearch ($N=16$, $T=8$)} & \textbf{0.2999$^{\dagger\ddagger}$} & \textbf{0.4960$^{\dagger\ddagger}$}	& \textbf{0.5362$^{\dagger\ddagger}$} & \textbf{0.4440$^{\dagger\ddagger}$} \\
        \bottomrule
    \end{tabular}}

    \label{tab:main-results}
    \vspace{-0.4cm}
\end{table*}

We observed that when the agent selects \textit{planning} as the first action, the subsequent steps consistently follow the outline of the generated plan. This suggests that encouraging multi-directional thinking about responding to the question can be achieved by generating a variety of plans, each exploring a different strategy for addressing the question. We study two methods for diverse planning:
\begin{itemize}[leftmargin=*]
    \item \textbf{Initial State Alteration:} If the initial state is modified before executing the planning action, the LLM can produce different plans, as the changes in personalized context influences the planning process. This can be modeled by varying the amount of past user interactions accessible to the agent, effectively addressing the challenge of long personal context. To implement this, we randomly select \( N \) subsets of the user profile \( P_u \), each of size \( \tau \times |P_u| \), where \( 0 \leq \tau \leq 1 \). Starting from these \( N \) diverse initial states lead to different planning trajectories and thinking pathways.
    
    \item \textbf{Planning Action Variation:} Another way to generate different plans from the same initial state is to use a higher sampling temperature \( 0 \leq \tau \leq 1 \) during plan generation, which increases the randomness in the generated plans, leading to more diverse outputs. Thus, we generate \( N \) different plans from the same initial state (created using the whole profile $P_u$) by setting a high temperature \( \tau \), leading to different thinking pathways.
\end{itemize}
Each thinking pathway results in a response, and together they form a set of response proposals \( R_{\text{PoT}} = \{r_i\}_{i=1}^{N} \). To produce the final response \( \hat{y}_{x_u} \) for the user, we need to ensure that it is the most suitable one for the user. To achieve this, we first use the LLM \( \pi \) to extract the important aspects \( I_u \) for user \( u \) from their user profile \( P_u \) using the prompt shown in Figure~\ref{fig:important-aspects-user-prompt} in Appendix~\ref{app:imp-details-response-agg}. This serves as an intermediate step to learn about the user's individual preferences. Based on this, there are two methods to generate the final response:
\begin{itemize}[leftmargin=*]

\item \textbf{Best-of-N:} Given the extracted important aspects \( I_u \) for the user $u$ and the input \( x_u \), the LLM \( \pi \) selects the best response from the set of responses resulted from multi-directional thinking \( R_{\text{PoT}} \), using the prompt shown in Figure~\ref{fig:best-of-N-prompt} in Appendix~\ref{app:imp-details-response-agg}.

\item \textbf{Mixture-of-N:} Given the extracted aspects \( I_u \) for the user \( u \) and the input \( x_u \), the LLM \( \pi \) combines different aspects of the generated responses from the set of multi-directional thinking outputs \( R_{\text{PoT}} \). The final response is denoted as \( \hat{y}_{x_u} = \pi(x_u, R_{\text{PoT}}, I_u) \), using the prompt shown in Figure~\ref{fig:mixture-of-N-prompt} in Appendix~\ref{app:imp-details-response-agg}. This mimics how humans solve problems by exploring multiple solutions and synthesizing a new solution that incorporates the strengths of them. This is beneficial for open-ended text generation, where there is no definitive right or wrong response. By aggregating the strengths of all responses, it generates a response that aligns with user preferences, ensuring a more personalized output. {\rmdpwithsearchshort} uses this approach as the primary response aggregation method.

\end{itemize}

\section{Experiments}

\subsection{Experimental Setup}

\subsubsection*{\textbf{Datasets}}

We conduct experiments on the LaMP-QA benchmark \citep{lamp-qa}, the only publicly available dataset for long-form personalized QA. LaMP-QA covers three diverse domains: (1) Art \& Entertainment (767 questions), (2) Lifestyle \& Personal Development (989 questions), and (3) Society \& Culture (1074 questions).
Each LaMP-QA example contains a user question, the user’s question history (serving as the user profile), a user-written narrative that reflects the user’s information need and intent, and a set of personalized rubrics specifying the aspects an ideal response should address. For evaluation, we follow the evaluation recipe of \citet{lamp-qa} with Gemini 1.5 Pro as the evaluator language model. For each question, the evaluator checks whether the generated response addresses each personalized aspect, assigning scores in the range $[0,2]$. These are normalized to $[0,1]$, and the final response score is computed as the mean normalized score across all aspects. For details, we refer the reader to \citet{lamp-qa}. In one experiment, we corroborate our findings through a side-by-side human evaluation comparing \rmdpwithsearchshort to the strongest baseline.

\subsubsection*{\textbf{\rmdpwithsearchshort's Configurations}}

We use Gemini 1.5 Pro as the backbone LLM. To show the generalizability of \rmdpwithsearchshort, we also use GPT-4o-mini\footnote{Available at: \url{https://platform.openai.com/docs/models/gpt-4o-mini}} \citep{openai2024gpt4ocard} using our most representative experiment configurations. For single pathways, we use Nucleus sampling \citep{Nucleus-sampling} with a temperature of \( \tau=0.1 \). For multi-directional pathways, we use the same setting, except for the \textit{Planning Action Variation}, where a temperature of \( \tau=0.9 \) is used exclusively for the \textit{planning} action. For generating diverse plans, we use \textit{Planning Action Variation} by default and \textit{Mixture-of-N} for response aggregation, unless otherwise specified. We set the context length to \( 32k \), the maximum number of actions per pathway to \( T=8 \) , and the number of pathways to \( N=16 \). In ablations, we reduce the number of pathways to \( N=4 \) to save costs, unless otherwise specified. For personalized context, we use the 10 recent questions of the user with the same order as they appear in the LaMP-QA \citep{lamp-qa} dataset.

\begin{figure*}
    \centering
    \includegraphics[width=0.85\textwidth]{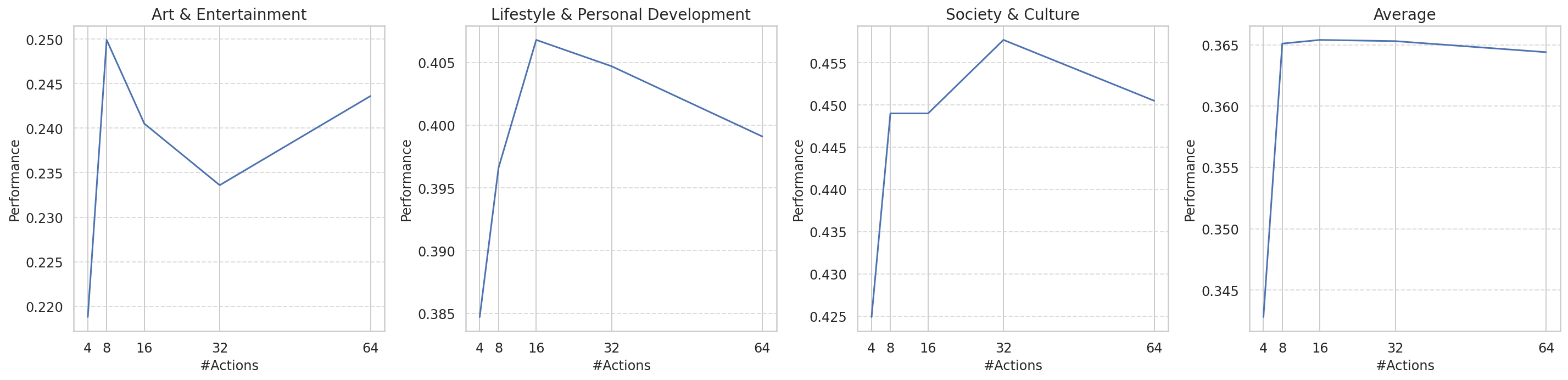}
    \vspace{-0.4cm}
    \caption{Effect of the maximum number of actions per pathway on the performance.}
    \label{fig:rmdp-steps-per-category}
\end{figure*}

\begin{figure*}
    \centering
    \vspace{-0.2cm}
    \includegraphics[width=0.85\textwidth]{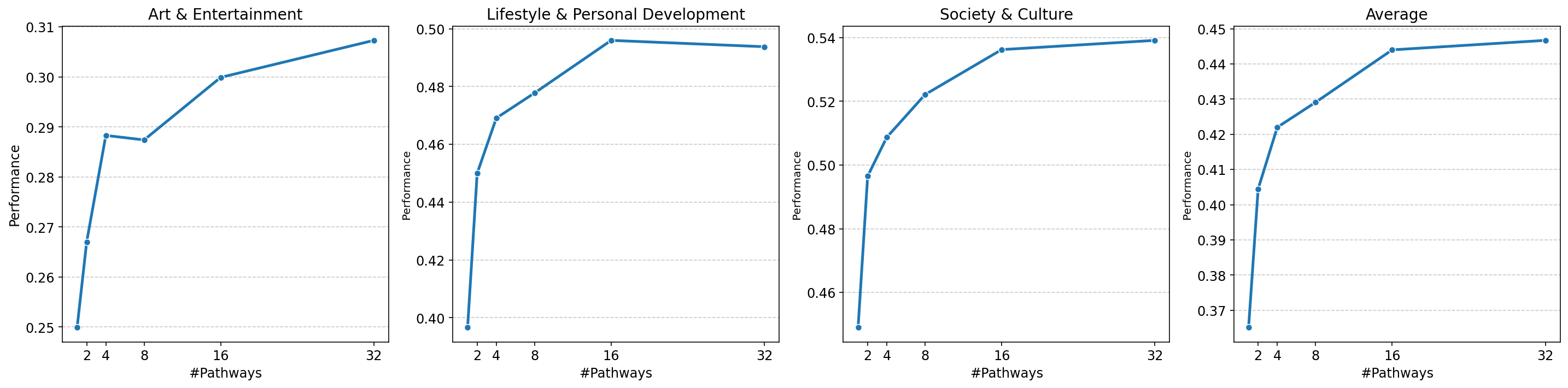}
    \vspace{-0.4cm}
    \caption{Effect of the number of pathways on the performance.}
    \label{fig:pathways-performance-per-category}
    \vspace{-0.4cm}
\end{figure*}

\subsubsection*{\textbf{{Baselines}}}

We use six baselines: (1) LLM without personalized context that directly answers the question, (2) LLM with access to the same personalized context as \rmdpwithsearchshort, and (3) LLM with access to personalized context that first performs reasoning over the context to extract important aspects relevant to the question before generating the response. The two latter use the prompts shown in Figures \ref{fig:baseline} and \ref{fig:baseline-cot} in Appendix \ref{app:baselines}. These baselines all use Nucleus sampling with a temperature of $\tau=0.1$. We extend the latter two baselines by doing multiple inferences and Best-of-N \citep{ichihara2025evaluation, brown2024largelanguagemonkeysscaling} and Mixture-of-N to evaluate their performance when test-time compute scales, to form baselines (5), (6), (7), and (8). Additionally, we include the (9) Tree of Thoughts (ToT) \citep{tot} as a baseline, leveraging its test-time compute scaling for comparison with our method. We follow their setup, particularly in the creative writing task, as it is the closest task they studied to personalized question answering. This setup consists of an intermediate planning step, where a set of plans is generated and the best plan is selected, followed by a final response generation step, where a set of responses is generated and the best response is chosen. For planning, we use the prompts shown in Figure~\ref{fig:plan-tot}, and for response generation, we refer to Figure~\ref{fig:gen-tot}, both in Appendix~\ref{app:baselines}. We observed that the baselines generate, on average, half the number of tokens compared to \({\rmdpwithsearchshort}\). To ensure a fair comparison, we set the Best-of-N and Mixture-of-N baselines to generate double the number of responses, with \(N = 32\) and a sampling temperature of $\tau=0.9$ to ensure diverse responses. For the Best-of-N and Mixture-of-N approach, we use the method introduced in Section \ref{sec:multi-directional}. For ToT, following \citet{tot} and their setup for the creative writing task that is the most similar to ours, we set the tree depth to \( B = 2 \). For fair comparison with \rmdpwithsearchshort, we generate \( N = 32 \) plans and responses at each node with a sampling temperature of $\tau=0.9$. The prompts for selecting the best plan is shown in Figure~\ref{fig:select-plan-tot} in Appendix~\ref{app:baselines}. For selecting the best response, we utilize the method introduced in Section \ref{sec:multi-directional}.

\subsection{Empirical Findings}
\label{sec:results}

\subsubsection*{\textbf{Comparison with baselines}} 

The results of this experiment are reported in Table~\ref{tab:main-results}. The findings show that \rmdpwithsearchshort outperforms all the baselines across all categories, achieving a \textbf{10.8\%} relative improvement on average over the best baseline (row 8). The improvements are statistically significant with respect to t-test with 95\% confidence.  This demonstrates the superiority of \rmdpwithsearchshort compared to the baselines. To compare our thinking-as-an-MDP method with baselines, we also report the results using only a single inference (rows 1-4). The results from this experiment show that our method (row 4) outperforms all single inference baselines (rows 1-3), with statistically significant improvements in two out of three categories and overall performance. Specifically, our method achieves a \textbf{7.2\%} relative improvement in average performance, suggesting that our thinking approach is significantly more effective than CoT.

\subsubsection*{\textbf{Effect of the maximum number of actions in \rmdpwithsearchshort}}

For this, we focus on a single inference ($N=1$) using \rmdpwithsearchshort and vary the maximum number of actions it can perform. The performance across all categories is shown in Figure~\ref{fig:rmdp-steps-per-category}. The findings show that increasing the number of actions improves performance up to a certain point (8 actions per pathway), after which performance declines. This occurs because, with too few actions, the LLM cannot construct effective pathways to solve the problem. Conversely, when given too many actions, the LLM may overthink the problem, performing unnecessary actions that degrade performance \cite{aggarwal2025optimalthinkingbenchevaluatingunderthinkingllms}.

\begin{figure*}
    \centering
    \includegraphics[width=0.85\textwidth]{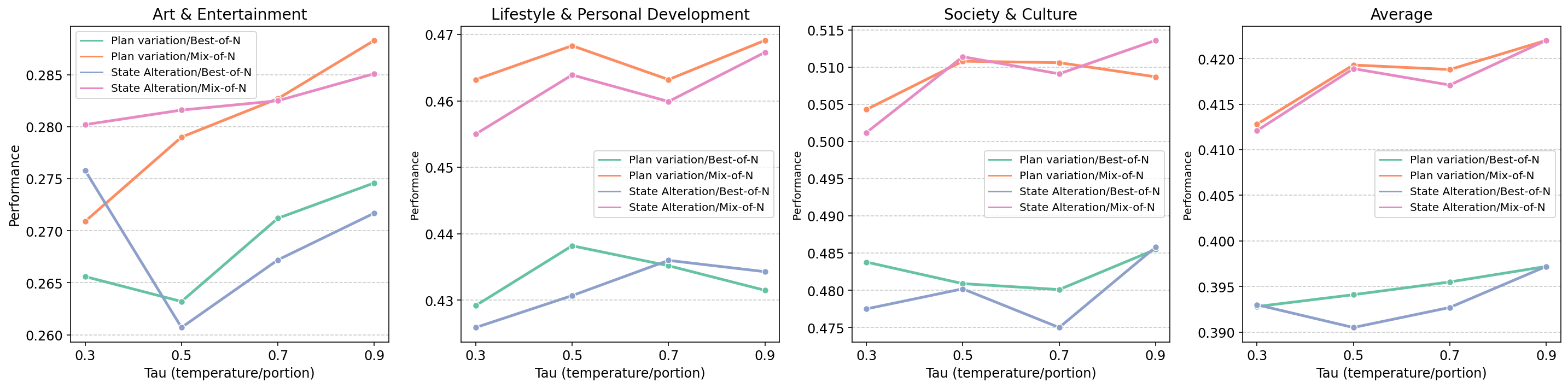}
    \vspace{-0.4cm}
    \caption{Effect of planning diversification and response aggregation methods on the performance.}
    \label{fig:thresh-select-combine-history-plan-per-category}
\end{figure*}

\begin{figure*}
    \centering
    \includegraphics[width=0.85\textwidth]{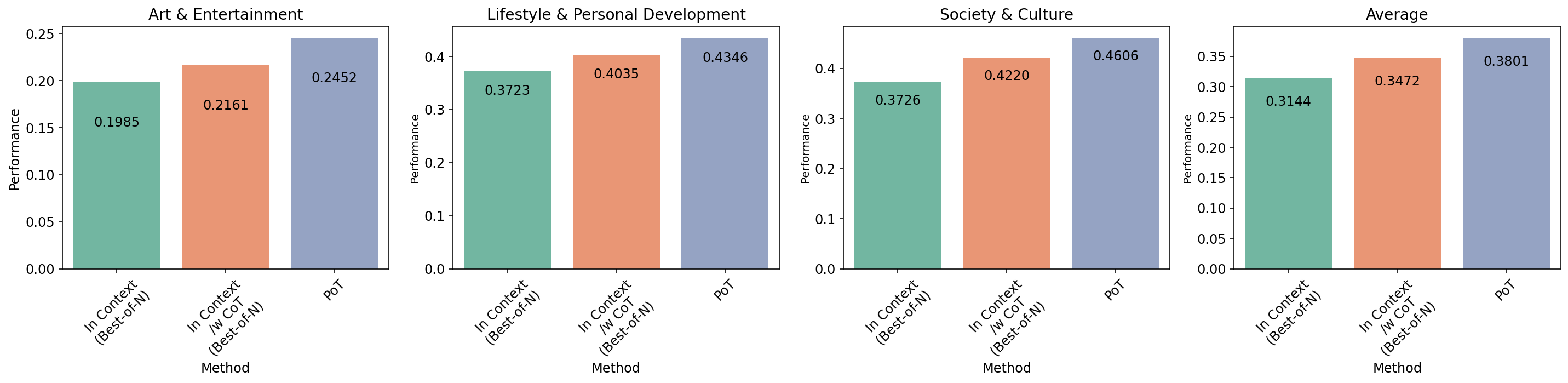}
    \vspace{-0.4cm}
    \caption{Performance of \rmdpwithsearch and baselines with GPT-4o-mini as the backbone LLM.}
    \label{fig:gpt4omini-per-category}
    \vspace{-0.2cm}
\end{figure*}


\subsubsection*{\textbf{Effect of the number of pathways in \rmdpwithsearchshort}}

We set the maximum number of actions per pathway to \( T=8 \) and vary the number of pathways. The performance across all categories is shown in Figure~\ref{fig:pathways-performance-per-category}. The findings indicate that increasing the number of pathways consistently improves performance; however, as the number grows, the improvements become thinner. This occurs because there is a limited set of different ways for answering the question, and after a certain point, the solutions start to repeat, leading to diminishing returns in performance with additional pathways.

\subsubsection*{\textbf{Effect of different plan diversification and response aggregation in \rmdpwithsearchshort}}

We evaluate all possible combinations of the methods introduced in Section~\ref{sec:multi-directional}. Specifically, we consider four configurations: 1) \textit{Planning Action Variation w/ Best-of-N}, 2) \textit{Planning Action Variation w/ Mixture-of-N}, 3) \textit{Initial State Alteration w/ Best-of-N}, and 4) \textit{Initial State Alteration w/ Mixture-of-N}. Additionally, we explore different values of the hyperparameter $\tau$ for both \textit{Planning Action Variation} and \textit{Initial State Alteration}. The results of this experiment are shown in Figure~\ref{fig:thresh-select-combine-history-plan-per-category}. Our analysis reveals several key observations. Most notably, \textit{Mixture-of-N} consistently outperforms \textit{Best-of-N} regardless of the plan diversification method. This can be attributed to the fact that \textit{Best-of-N} selects only one response from the generated candidates, whereas \textit{Mixture-of-N} enables the combination of strengths from multiple responses, resulting in a more comprehensive and higher-quality output. we observe that increasing the hyperparameter $\tau$ improves performance. In \textit{Planning Action Variation}, a higher $\tau$ leads to a more diverse set of generated plans due to higher sampling temperature, resulting in more varied thinking pathways that may produce more tailored responses. In \textit{Initial State Alteration}, increasing $\tau$ provides each pathway with more user-specific information, enabling more personalized responses. Comparing the two, \textit{Planning Action Variation} generally outperforms \textit{Initial State Alteration} in terms of average performance. However, as $\tau$ increases, the performance gap narrows. Overall, \textit{Planning Action Variation w/ Mixture-of-N} achieves the best performance for personalized question answering.

\subsubsection*{\textbf{\rmdpwithsearchshort's generalizability and independence of LLM}}

In the same setting as Gemini 1.5 pro, we apply \rmdpwithsearchshort and the best performing baselines to another backbone LLM, GPT-4o-mini, to demonstrate its effectiveness across different LLMs with long context windows. The average performance across all categories is presented in Figure~\ref{fig:gpt4omini-per-category}. The results indicate that, similar to Gemini 1.5 Pro, GPT-4o-mini with \rmdpwithsearchshort outperforms all baselines on all categories. It achieves a statistically significant 9.4\% relative improvement on average across all datasets compared to the best baseline. These findings suggest that \rmdpwithsearchshort enhances the performance of LLMs in personalized QA regardless of the underlying backbone LLM.

\subsubsection*{\textbf{{Human evaluation.}}}

We randomly sample 100 examples from \datasetshort and generate outputs using \rmdpwithsearchshort and {In-Context Personalization w/ CoT \& Best-of-32}. Human annotators evaluate the outputs based on their alignment with the the user's question narrative (i.e., a detailed description of user's information need and perspective). Each example is annotated twice, yielding an inter-annotator agreement Cohen's kappa of $0.613$. The results indicate that annotators preferred \rmdpwithsearchshort in \textbf{66\%} of the cases, the baseline in \textbf{19\%}, and reported tie in \textbf{15\%} of cases. These results show that \rmdpwithsearchshort achieves considerably higher alignment with human preferences.

\subsection{Case Study}

We observe that \rmdpwithsearchshort($N=16$, $T=8$) generates $1943$ unique pathways, defined by distinct action sequences chosen. Specifically, \textit{Art \& Entertainment} yields $1109$, \textit{Lifestyle \& Personal Development} produces $814$, and \textit{Society \& Culture} results in $1182$ unique pathways. On average, each example generates $11.1$ unique pathways with on average $7.19$ actions. As a case study demonstrating the effectiveness of \rmdpwithsearchshort, we present an example in Figure~\ref{fig:case-study} in Appendix:

\begin{itemize}[leftmargin=*]

\item{\textbf{Thinking Pathways \& Generated Responses:}} To conserve space, we present in Figure~\ref{fig:case-study} only the sequence of steps taken in each pathway, rather than the full generated text, and the response in each pathway. Each pathway corresponds to a distinct reasoning trajectory generated by \rmdpwithsearchshort. For example, in the illustrated case, \rmdpwithsearchshort selects different pathways depending on whether personalization is required. Some pathways generate a direct response without personalization (e.g., Pathway 7), while others involve multiple rounds of reasoning that integrate various aspects of the query with the user profile before producing an answer (Pathways 1–6, 9–11, 13–15). Certain pathways also involve revision (Pathways 12, 15, and 16) or clarification (Pathway 4). Notably, we observed that for some trajectories \rmdpwithsearchshort omit personalization entirely (Paths 7, 8, and 11). This shows that \rmdpwithsearchshort deliberately explores a diverse range of reasoning strategies, producing responses from multiple perspectives that are subsequently combined into a final answer encompassing the advantages of each. Although responses across pathways share common elements, each retains distinctive characteristics, reflecting the variation in underlying reasoning strategies. 

\item{\textbf{Final Response:}} Figure~\ref{fig:case-study} shows the final response to the user’s question, produced by aggregating and integrating components from multiple pathway outputs into a single coherent answer. The contributions of different pathways are indicated in brackets ([]). This shows how \rmdpwithsearchshort leverages the complementary strengths of diverse reasonings, combining them into a unified response that aligns with the user’s preferences and key information needs.

\end{itemize}

\section{Conclusions}

This paper addresses the critical yet under-explored problem of personalized question answering. we propose \textit{\rmdpwithsearch (\rmdpwithsearchshort)}, a novel approach that models the LLM's thinking process as an MDP during test time. By enabling the exploration of multiple thinking pathways and aggregating diverse responses, \textit{\rmdpwithsearchshort} effectively captures user preferences to generate personalized answers. Our results on \textit{\datasetshort} demonstrate that \textit{\rmdpwithsearchshort} achieves substantial improvements over baseline methods, with performance gains of up to 10.8\%. Furthermore, human evaluation confirms the effectiveness of our approach, showing a strong user preference over existing baselines. This paper focuses solely on personalized question answering; however \rmdpwithsearch is a general framework and future work can explore the potential of \rmdpwithsearch for long-form text generation beyond personalization.

\section*{Acknowledgement}

This work was supported in part by the Center for Intelligent Information Retrieval, in part by NSF grant \#2143434, in part by the Office of Naval Research contract \#N000142412612, and with support from by Google.org. Any opinions, findings and conclusions or recommendations expressed in this material are those of the authors and do not necessarily reflect those of the sponsor.

\bibliographystyle{ACM-Reference-Format}
\bibliography{sample-base}

@inproceedings{se-pqa,
author = {Kasela, Pranav and Braga, Marco and Pasi, Gabriella and Perego, Raffaele},
title = {SE-PQA: Personalized Community Question Answering},
year = {2024},
isbn = {9798400701726},
publisher = {Association for Computing Machinery},
address = {New York, NY, USA},
url = {https://doi.org/10.1145/3589335.3651445},
doi = {10.1145/3589335.3651445},
booktitle = {Companion Proceedings of the ACM Web Conference 2024},
pages = {1095–1098},
numpages = {4},
location = {Singapore, Singapore},
series = {WWW '24}
}

@article{thinking-process-1,
author = {Frensch, Peter and Funke, Joachim},
year = {2002},
month = {01},
pages = {},
title = {Thinking and problem solving},
journal = {Encyclopedia of Life Support Systems (EOLSS), Developed under the Auspices of the UNESCO, Eolss Publishers, Oxford ,UK}
}

@inproceedings{search-r1,
title={Search-R1: Training {LLM}s to Reason and Leverage Search Engines with Reinforcement Learning},
author={Bowen Jin and Hansi Zeng and Zhenrui Yue and Jinsung Yoon and Sercan O Arik and Dong Wang and Hamed Zamani and Jiawei Han},
booktitle={Second Conference on Language Modeling},
year={2025},
url={https://openreview.net/forum?id=Rwhi91ideu}
}

@inproceedings{thinking-process-2,
  title={Complex Cognition: The Psychology of Human Thought},
  author={Robert J. Sternberg and Talia Ben-Zeev},
  year={2001}
}

@misc{salemi2025reasoningenhancedselftraininglongformpersonalized,
      title={Reasoning-Enhanced Self-Training for Long-Form Personalized Text Generation}, 
      author={Alireza Salemi and Cheng Li and Mingyang Zhang and Qiaozhu Mei and Weize Kong and Tao Chen and Zhuowan Li and Michael Bendersky and Hamed Zamani},
      year={2025},
      eprint={2501.04167},
      archivePrefix={arXiv},
      primaryClass={cs.CL},
      url={https://arxiv.org/abs/2501.04167}, 
}

@inproceedings{cot,
author = {Wei, Jason and Wang, Xuezhi and Schuurmans, Dale and Bosma, Maarten and Ichter, Brian and Xia, Fei and Chi, Ed H. and Le, Quoc V. and Zhou, Denny},
title = {Chain-of-thought prompting elicits reasoning in large language models},
year = {2022},
isbn = {9781713871088},
publisher = {Curran Associates Inc.},
address = {Red Hook, NY, USA},
booktitle = {Proceedings of the 36th International Conference on Neural Information Processing Systems},
articleno = {1800},
numpages = {14},
location = {New Orleans, LA, USA},
series = {NIPS '22}
}

@inproceedings{
liu2023logicot,
title={LogiCoT: Logical Chain-of-Thought Instruction Tuning},
author={Hanmeng Liu and Zhiyang Teng and Leyang Cui and Chaoli Zhang and Qiji Zhou and Yue Zhang},
booktitle={The 2023 Conference on Empirical Methods in Natural Language Processing},
year={2023},
url={https://openreview.net/forum?id=qlCtkvgQJH}
}

@InProceedings{10.1007/978-981-97-7232-2_13,
author="Yin, Han
and Yu, Jianxing
and Lin, Miaopei
and Wang, Shiqi",
editor="Zhang, Wenjie
and Tung, Anthony
and Zheng, Zhonglong
and Yang, Zhengyi
and Wang, Xiaoyang
and Guo, Hongjie",
title="Answering Spatial Commonsense Questions Based on Chain-of-Thought Reasoning with Adaptive Complexity",
booktitle="Web and Big Data",
year="2024",
publisher="Springer Nature Singapore",
address="Singapore",
pages="186--200",
isbn="978-981-97-7232-2"
}

@misc{openai_o1,
  title = {OpenAI o1 System Card},
  author = {{OpenAI}},
  year = {2024}
}

@misc{deepseek-r1,
      title={DeepSeek-R1: Incentivizing Reasoning Capability in LLMs via Reinforcement Learning}, 
      author={DeepSeek-AI},
      year={2025},
      eprint={2501.12948},
      archivePrefix={arXiv},
      primaryClass={cs.CL} 
}

@inproceedings{
tot,
title={Tree of Thoughts: Deliberate Problem Solving with Large Language Models},
author={Shunyu Yao and Dian Yu and Jeffrey Zhao and Izhak Shafran and Thomas L. Griffiths and Yuan Cao and Karthik R Narasimhan},
booktitle={Thirty-seventh Conference on Neural Information Processing Systems},
year={2023},
url={https://openreview.net/forum?id=5Xc1ecxO1h}
}

@inproceedings{multi-agent,
  title     = {Large Language Model Based Multi-agents: A Survey of Progress and Challenges},
  author    = {Guo, Taicheng and Chen, Xiuying and Wang, Yaqi and Chang, Ruidi and Pei, Shichao and Chawla, Nitesh V. and Wiest, Olaf and Zhang, Xiangliang},
  booktitle = {Proceedings of the Thirty-Third International Joint Conference on
               Artificial Intelligence, {IJCAI-24}},
  publisher = {International Joint Conferences on Artificial Intelligence Organization},
  editor    = {Kate Larson},
  pages     = {8048--8057},
  year      = {2024},
  month     = {8},
  note      = {Survey Track},
  doi       = {10.24963/ijcai.2024/890},
  url       = {https://doi.org/10.24963/ijcai.2024/890},
}

@inproceedings{
hao2023reasoning,
title={Reasoning with Language Model is Planning with World Model},
author={Shibo Hao and Yi Gu and Haodi Ma and Joshua Jiahua Hong and Zhen Wang and Daisy Zhe Wang and Zhiting Hu},
booktitle={The 2023 Conference on Empirical Methods in Natural Language Processing},
year={2023},
url={https://openreview.net/forum?id=VTWWvYtF1R}
}

@article{yao2022react,
  title={ReAct: Synergizing Reasoning and Acting in Language Models},
  author={Yao, Shunyu and Zhao, Jeffrey and Yu, Dian and Du, Nan and Shafran, Izhak and Narasimhan, Karthik and Cao, Yuan},
  journal={arXiv preprint arXiv:2210.03629},
  year={2022}
}

@inproceedings{
fu2024autoguide,
title={AutoGuide: Automated Generation and Selection of Context-Aware Guidelines for Large Language Model Agents},
author={Yao Fu and Dong-Ki Kim and Jaekyeom Kim and Sungryull Sohn and Lajanugen Logeswaran and Kyunghoon Bae and Honglak Lee},
booktitle={The Thirty-eighth Annual Conference on Neural Information Processing Systems},
year={2024},
url={https://openreview.net/forum?id=mRIQz8Zd6O}
}

@inproceedings{
xie2024a,
title={A Human-Like Reasoning Framework for Multi-Phases Planning Task with Large Language Models},
author={Chengxing Xie and Difan Zou},
booktitle={ICML 2024 Workshop on LLMs and Cognition},
year={2024},
url={https://openreview.net/forum?id=BX6wC2452j}
}

@inproceedings{
liu2025toolplanner,
title={Tool-Planner: Task Planning with Clusters across Multiple Tools},
author={Yanming Liu and Xinyue Peng and Jiannan Cao and Shi Bo and Yuwei Zhang and Xuhong Zhang and Sheng Cheng and Xun Wang and Jianwei Yin and Tianyu Du},
booktitle={The Thirteenth International Conference on Learning Representations},
year={2025},
url={https://openreview.net/forum?id=dRz3cizftU}
}

@misc{park2024ensemblinglargelanguagemodels,
      title={Ensembling Large Language Models with Process Reward-Guided Tree Search for Better Complex Reasoning}, 
      author={Sungjin Park and Xiao Liu and Yeyun Gong and Edward Choi},
      year={2024},
      eprint={2412.15797},
      archivePrefix={arXiv},
      primaryClass={cs.CL}
}

@misc{wang2024qimprovingmultistepreasoning,
      title={Q*: Improving Multi-step Reasoning for LLMs with Deliberative Planning}, 
      author={Chaojie Wang and Yanchen Deng and Zhiyi Lyu and Liang Zeng and Jujie He and Shuicheng Yan and Bo An},
      year={2024},
      eprint={2406.14283},
      archivePrefix={arXiv},
      primaryClass={cs.AI}
}

@article{
ichihara2025evaluation,
title={Evaluation of Best-of-N Sampling Strategies for Language Model Alignment},
author={Yuki Ichihara and Yuu Jinnai and Tetsuro Morimura and Kenshi Abe and Kaito Ariu and Mitsuki Sakamoto and Eiji Uchibe},
journal={Transactions on Machine Learning Research},
issn={2835-8856},
year={2025},
url={https://openreview.net/forum?id=H4S4ETc8c9},
note={}
}

@inproceedings{
Nucleus-sampling,
title={The Curious Case of Neural Text Degeneration},
author={Ari Holtzman and Jan Buys and Li Du and Maxwell Forbes and Yejin Choi},
booktitle={International Conference on Learning Representations},
year={2020},
url={https://openreview.net/forum?id=rygGQyrFvH}
}

@misc{openai2024gpt4ocard,
      title={GPT-4o System Card}, 
      author={OpenAI},
      year={2024},
      eprint={2410.21276},
      archivePrefix={arXiv},
      primaryClass={cs.CL}
}

@inproceedings{10.1145/2702123.2702503,
author = {Fowler, Andrew and Partridge, Kurt and Chelba, Ciprian and Bi, Xiaojun and Ouyang, Tom and Zhai, Shumin},
title = {Effects of Language Modeling and Its Personalization on Touchscreen Typing Performance},
year = {2015},
isbn = {9781450331456},
publisher = {Association for Computing Machinery},
address = {New York, NY, USA},
url = {https://doi.org/10.1145/2702123.2702503},
doi = {10.1145/2702123.2702503},
booktitle = {Proceedings of the 33rd Annual ACM Conference on Human Factors in Computing Systems},
pages = {649–658},
numpages = {10},
location = {Seoul, Republic of Korea},
series = {CHI '15}
}

@article{10.1145/1462198.1462203,
author = {Xue, Gui-Rong and Han, Jie and Yu, Yong and Yang, Qiang},
title = {User Language Model for Collaborative Personalized Search},
year = {2009},
issue_date = {February 2009},
publisher = {Association for Computing Machinery},
address = {New York, NY, USA},
volume = {27},
number = {2},
issn = {1046-8188},
url = {https://doi.org/10.1145/1462198.1462203},
doi = {10.1145/1462198.1462203},
journal = {ACM Trans. Inf. Syst.},
month = {mar},
articleno = {11},
numpages = {28}
}

@misc{naumov2019deep,
      title={Deep Learning Recommendation Model for Personalization and Recommendation Systems}, 
      author={Maxim Naumov and Dheevatsa Mudigere and Hao-Jun Michael Shi and Jianyu Huang and Narayanan Sundaraman and Jongsoo Park and Xiaodong Wang and Udit Gupta and Carole-Jean Wu and Alisson G. Azzolini and Dmytro Dzhulgakov and Andrey Mallevich and Ilia Cherniavskii and Yinghai Lu and Raghuraman Krishnamoorthi and Ansha Yu and Volodymyr Kondratenko and Stephanie Pereira and Xianjie Chen and Wenlin Chen and Vijay Rao and Bill Jia and Liang Xiong and Misha Smelyanskiy},
      year={2019},
      eprint={1906.00091},
      archivePrefix={arXiv},
      primaryClass={cs.IR}
}

@misc{li2023teach,
      title={Teach LLMs to Personalize -- An Approach inspired by Writing Education}, 
      author={Cheng Li and Mingyang Zhang and Qiaozhu Mei and Yaqing Wang and Spurthi Amba Hombaiah and Yi Liang and Michael Bendersky},
      year={2023},
      eprint={2308.07968},
      archivePrefix={arXiv}
}

@misc{mysore2023pearl,
      title={PEARL: Personalizing Large Language Model Writing Assistants with Generation-Calibrated Retrievers}, 
      author={Sheshera Mysore and Zhuoran Lu and Mengting Wan and Longqi Yang and Steve Menezes and Tina Baghaee and Emmanuel Barajas Gonzalez and Jennifer Neville and Tara Safavi},
      year={2023},
      eprint={2311.09180},
      archivePrefix={arXiv}
}

@inproceedings{zhang-etal-2024-llm-based,
    title = "{LLM}-based Medical Assistant Personalization with Short- and Long-Term Memory Coordination",
    author = "Zhang, Kai  and
      Kang, Yangyang  and
      Zhao, Fubang  and
      Liu, Xiaozhong",
    editor = "Duh, Kevin  and
      Gomez, Helena  and
      Bethard, Steven",
    booktitle = "Proceedings of the 2024 Conference of the North American Chapter of the Association for Computational Linguistics: Human Language Technologies (Volume 1: Long Papers)",
    month = jun,
    year = "2024",
    address = "Mexico City, Mexico",
    publisher = "Association for Computational Linguistics",
    url = "https://aclanthology.org/2024.naacl-long.132",
    pages = "2386--2398",
}

@misc{lu2024corporate,
      title={Corporate Communication Companion (CCC): An LLM-empowered Writing Assistant for Workplace Social Media}, 
      author={Zhuoran Lu and Sheshera Mysore and Tara Safavi and Jennifer Neville and Longqi Yang and Mengting Wan},
      year={2024},
      eprint={2405.04656},
      archivePrefix={arXiv}
}

@misc{peft-rag-personalization,
      title={Comparing Retrieval-Augmentation and Parameter-Efficient Fine-Tuning for Privacy-Preserving Personalization of Large Language Models}, 
      author={Alireza Salemi and Hamed Zamani},
      year={2024},
      eprint={2409.09510},
      archivePrefix={arXiv},
      primaryClass={cs.CL} 
}

@misc{jang2023personalized,
      title={Personalized Soups: Personalized Large Language Model Alignment via Post-hoc Parameter Merging}, 
      author={Joel Jang and Seungone Kim and Bill Yuchen Lin and Yizhong Wang and Jack Hessel and Luke Zettlemoyer and Hannaneh Hajishirzi and Yejin Choi and Prithviraj Ammanabrolu},
      year={2023},
      eprint={2310.11564},
      archivePrefix={arXiv}
}

@inproceedings{Li_2024, series={WWW ’24},
   title={Learning to Rewrite Prompts for Personalized Text Generation},
   url={http://dx.doi.org/10.1145/3589334.3645408},
   DOI={10.1145/3589334.3645408},
   booktitle={Proceedings of the ACM on Web Conference 2024},
   publisher={ACM},
   author={Li, Cheng and Zhang, Mingyang and Mei, Qiaozhu and Kong, Weize and Bendersky, Michael},
   year={2024},
   month=may, collection={WWW ’24} }

@misc{tan2024personalized,
      title={Personalized Pieces: Efficient Personalized Large Language Models through Collaborative Efforts}, 
      author={Zhaoxuan Tan and Zheyuan Liu and Meng Jiang},
      year={2024},
      eprint={2406.10471},
      archivePrefix={arXiv}
}

@inproceedings{lamp,
    title = "{L}a{MP}: When Large Language Models Meet Personalization",
    author = "Salemi, Alireza  and
      Mysore, Sheshera  and
      Bendersky, Michael  and
      Zamani, Hamed",
    editor = "Ku, Lun-Wei  and
      Martins, Andre  and
      Srikumar, Vivek",
    booktitle = "Proceedings of the 62nd Annual Meeting of the Association for Computational Linguistics (Volume 1: Long Papers)",
    month = aug,
    year = "2024",
    address = "Bangkok, Thailand",
    publisher = "Association for Computational Linguistics",
    url = "https://aclanthology.org/2024.acl-long.399",
    pages = "7370--7392",
}

@inproceedings{rspg,
author = {Salemi, Alireza and Kallumadi, Surya and Zamani, Hamed},
title = {Optimization Methods for Personalizing Large Language Models through Retrieval Augmentation},
year = {2024},
isbn = {9798400704314},
publisher = {Association for Computing Machinery},
address = {New York, NY, USA},
url = {https://doi.org/10.1145/3626772.3657783},
doi = {10.1145/3626772.3657783},
booktitle = {Proceedings of the 47th International ACM SIGIR Conference on Research and Development in Information Retrieval},
pages = {752–762},
numpages = {11},
location = {Washington DC, USA},
series = {SIGIR '24}
}

@misc{aggarwal2025optimalthinkingbenchevaluatingunderthinkingllms,
      title={OptimalThinkingBench: Evaluating Over and Underthinking in LLMs}, 
      author={Pranjal Aggarwal and Seungone Kim and Jack Lanchantin and Sean Welleck and Jason Weston and Ilia Kulikov and Swarnadeep Saha},
      year={2025},
      eprint={2508.13141},
      archivePrefix={arXiv},
      primaryClass={cs.CL},
      url={https://arxiv.org/abs/2508.13141}, 
}

@misc{longlamp,
      title={LongLaMP: A Benchmark for Personalized Long-form Text Generation}, 
      author={Ishita Kumar and Snigdha Viswanathan and Sushrita Yerra and Alireza Salemi and Ryan A. Rossi and Franck Dernoncourt and Hanieh Deilamsalehy and Xiang Chen and Ruiyi Zhang and Shubham Agarwal and Nedim Lipka and Hamed Zamani},
      year={2024},
      eprint={2407.11016},
      archivePrefix={arXiv},
      primaryClass={cs.CL} 
}

@misc{snell2024scalingllmtesttimecompute,
      title={Scaling LLM Test-Time Compute Optimally can be More Effective than Scaling Model Parameters}, 
      author={Charlie Snell and Jaehoon Lee and Kelvin Xu and Aviral Kumar},
      year={2024},
      eprint={2408.03314},
      archivePrefix={arXiv},
      primaryClass={cs.LG}
}

@misc{lamp-qa,
      title={LaMP-QA: A Benchmark for Personalized Long-form Question Answering}, 
      author={Alireza Salemi and Hamed Zamani},
      year={2025},
      eprint={2506.00137},
      archivePrefix={arXiv},
      primaryClass={cs.CL}, 
}

@misc{salemi2025learningnaturallanguagefeedback,
      title={Learning from Natural Language Feedback for Personalized Question Answering}, 
      author={Alireza Salemi and Hamed Zamani},
      year={2025},
      eprint={2508.10695},
      archivePrefix={arXiv},
      primaryClass={cs.CL},
      url={https://arxiv.org/abs/2508.10695}, 
}

@misc{chen2024simpleprovablescalinglaw,
      title={A Simple and Provable Scaling Law for the Test-Time Compute of Large Language Models}, 
      author={Yanxi Chen and Xuchen Pan and Yaliang Li and Bolin Ding and Jingren Zhou},
      year={2024},
      eprint={2411.19477},
      archivePrefix={arXiv},
      primaryClass={cs.CL}
}

@misc{bi2024forestofthoughtscalingtesttimecompute,
      title={Forest-of-Thought: Scaling Test-Time Compute for Enhancing LLM Reasoning}, 
      author={Zhenni Bi and Kai Han and Chuanjian Liu and Yehui Tang and Yunhe Wang},
      year={2024},
      eprint={2412.09078},
      archivePrefix={arXiv},
      primaryClass={cs.CL} 
}

@misc{zhang2024scalingllminferenceoptimized,
      title={Scaling LLM Inference with Optimized Sample Compute Allocation}, 
      author={Kexun Zhang and Shang Zhou and Danqing Wang and William Yang Wang and Lei Li},
      year={2024},
      eprint={2410.22480},
      archivePrefix={arXiv},
      primaryClass={cs.CL}
}

@misc{brown2024largelanguagemonkeysscaling,
      title={Large Language Monkeys: Scaling Inference Compute with Repeated Sampling}, 
      author={Bradley Brown and Jordan Juravsky and Ryan Ehrlich and Ronald Clark and Quoc V. Le and Christopher Ré and Azalia Mirhoseini},
      year={2024},
      eprint={2407.21787},
      archivePrefix={arXiv},
      primaryClass={cs.LG}
}

\appendix

\begin{figure}
    \centering
    \includegraphics[width=0.9\linewidth]{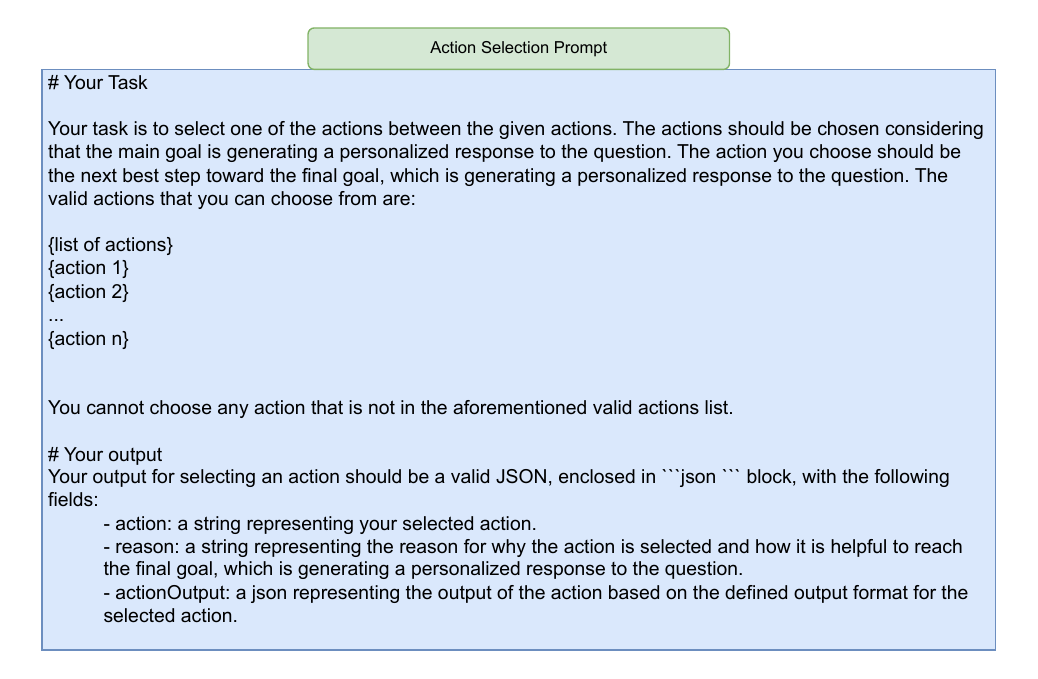}
    \vspace{-0.6cm}
    \caption{Action selection prompt used in \rmdpwithsearchshort.}
    \label{fig:action-selection-prompt}
    \vspace{-0.5cm}
\end{figure}

\begin{figure}
    \centering
    \includegraphics[width=0.9\linewidth]{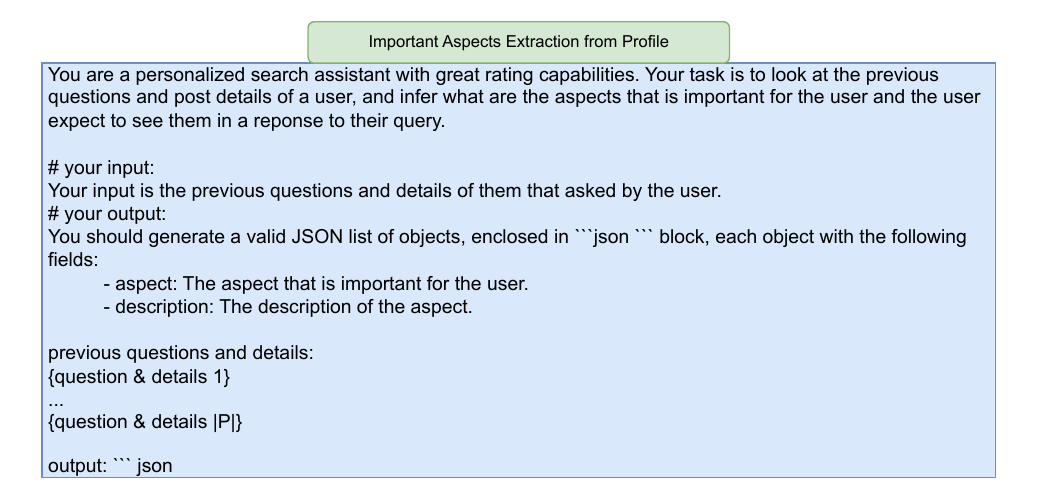}
    \vspace{-0.6cm}
    \caption{Prompt for extracting important aspects from the user profile used in \rmdpwithsearchshort.}
    \label{fig:important-aspects-user-prompt}
    \vspace{-0.5cm}
\end{figure}

\begin{figure}
    \centering
    \includegraphics[width=0.9\linewidth]{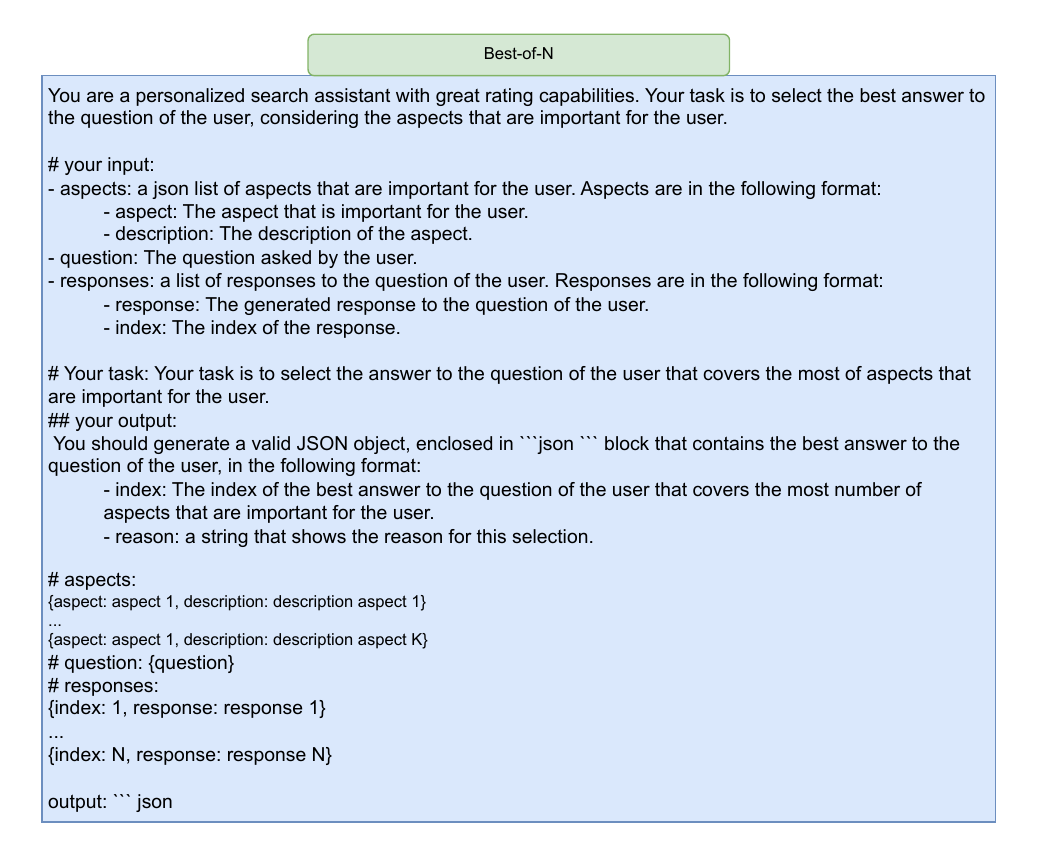}
    \vspace{-0.6cm}
    \caption{Prompt for selecting the best response out of N generated responses used in \rmdpwithsearchshort.}
    \label{fig:best-of-N-prompt}
    \vspace{-0.5cm}
\end{figure}

\begin{figure}
    \centering
    \includegraphics[width=0.9\linewidth]{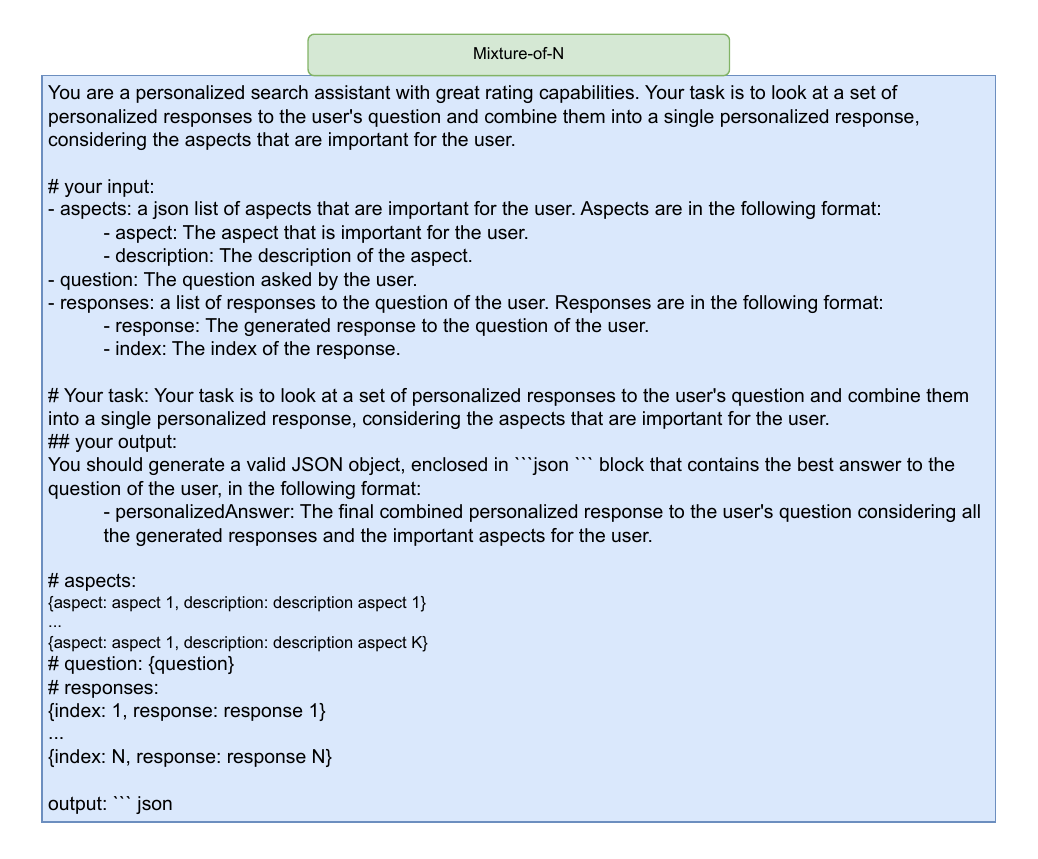}
    \vspace{-0.6cm}
    \caption{Prompt for mixing the N generated responses into a single response used in \rmdpwithsearchshort.}
    \label{fig:mixture-of-N-prompt}
    \vspace{-0.8cm}
\end{figure}


\section{\rmdpwithsearchshort's Implementation Details}
\label{app:imp-details}

\subsection{Actions Definition}
\label{app:imp-details-action-def}

We define an action as a means by which the agent interacts with the text generation environment, specifically the LLM, by instructing it to perform a particular operation. Each action is associated with a specific output format that the LLM generates in response to the instruction. In this paper, we consider the following actions as the operations the model can take to respond to a question:
\begin{itemize}[leftmargin=*]
    \item \textit{answering:} This action instructs the model to generate a personalized response to the question, provided that personalization can enhance the response. It should be selected when the model is prepared to answer the question, ensuring that all necessary information is available and there is no ambiguity.
    
    \item \textit{planning:} This action instructs the model to generate a plan of steps required to address all aspects of the question. It should be executed when the question has multiple components, and the model recognizes that planning the next steps is necessary to adequately respond to the user's query.
    
    \item \textit{personalization:} This action asks the model to evaluate whether the user's question can benefit from personalization. It is useful when determining whether personalizing the response is necessary or if a generic response would suffice.
    
    \item \textit{personalizing:} This action asks the model to identify relevant information from the user profile and summarize how it can be used to answer the question. It is executed when personalized information is necessary to enhance the quality of the response.
    
    \item \textit{reasoning:} This action prompts the model to break down one aspect of the question and think through it step by step. It is executed when questions involve multiple aspects, and the model needs to focus on and consider how to address a specific aspect.
    
    \item \textit{clarifying:} This action prompts the model to assess whether the question is ambiguous and determine how it can be clarified. It is executed when the model identifies that the current information is insufficient or unclear, requiring further elaboration.
    
    \item \textit{summarizing:} This action instructs the model to summarize all available information and findings. It should be executed when the model determines that the current information needs to be consolidated or synthesized to ensure a clearer understanding before continuing with the response generation.
    
    \item \textit{revising:} This action instructs the model to revise the current response to the question. It should be executed when the model has already generated an initial response but believes the response can be improved or adjusted to better address the user's query or to incorporate additional insights or information.
    
    \item \textit{finalizing:} This action finalizes the response. It should be executed when the model believes the response is complete, accurate, and addresses the user's query. No further actions or revisions are necessary, and the response is ready to be presented. 
    
\end{itemize}

\begin{figure}
    \centering
    \includegraphics[width=0.9\linewidth]{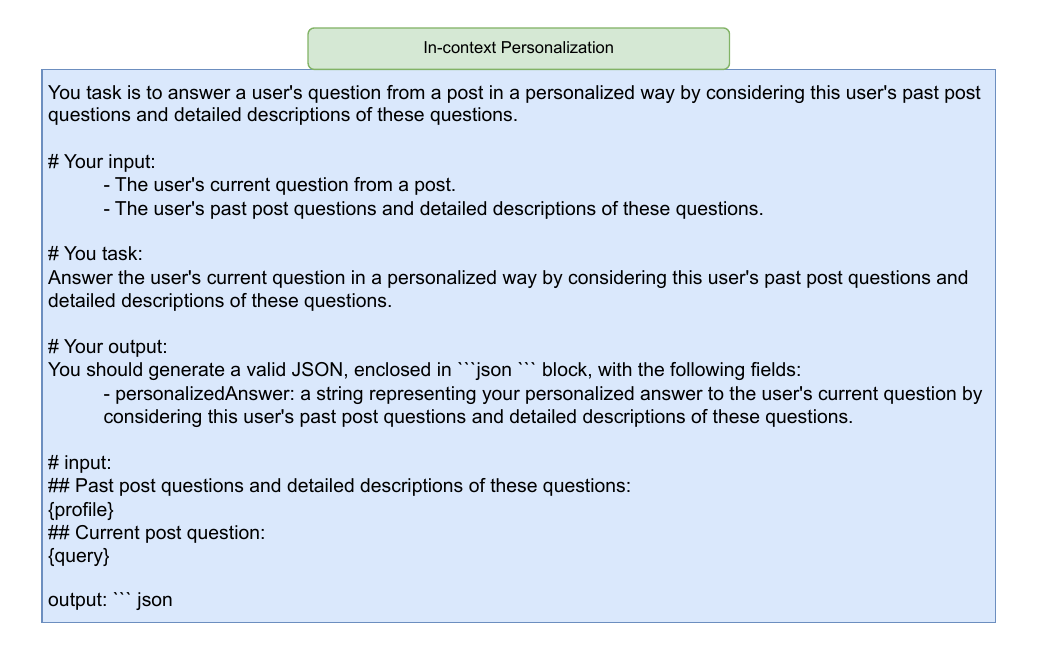}
    \vspace{-0.6cm}
    \caption{Prompt for the In-Context Personalization.}
    \label{fig:baseline}
\end{figure}

\begin{figure}
    \centering
    \includegraphics[width=0.9\linewidth]{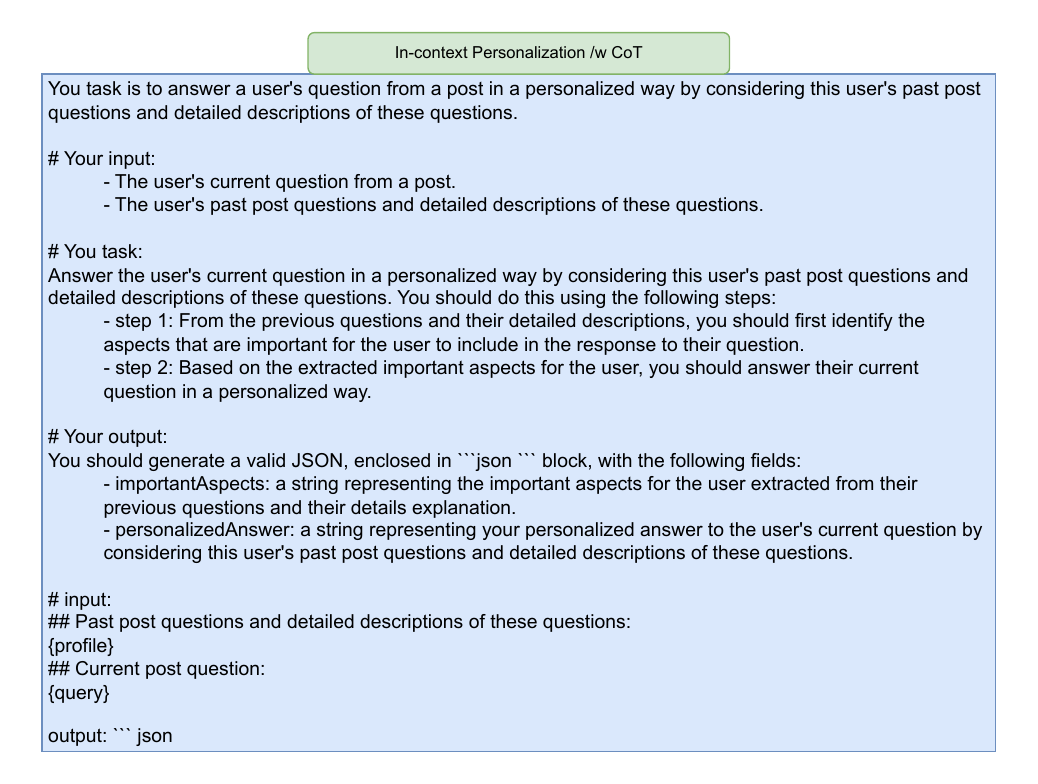}
    \vspace{-0.6cm}
    \caption{Prompt for In-Context Personalization with CoT.}
    \label{fig:baseline-cot}
\end{figure}

\subsection{Details of Response Aggregation Methods}
\label{app:imp-details-response-agg}

An essential component of \rmdpwithsearch involves aggregating the candidate responses generated across different thinking pathways into a single output. Each pathway produces a candidate response, collectively forming a set of response proposals, denoted as \( R_{\text{PoT}} \). To generate the final response \( \hat{y}_{x_u} \) personalized for the user, it is crucial to select the most suitable candidate. To achieve this, we first employ the LLM \( \pi \) to extract the key aspects \( I_u \) relevant to user \( u \) from their user profile \( P_u \). This extraction process is guided by the prompt illustrated in Figure~\ref{fig:important-aspects-user-prompt}, acting as an intermediate step to capture the user's individual preferences. Based on this, there are two methods to generate the final response:
\begin{itemize}[leftmargin=*]
    \item \textit{Best-of-N:} Given the extracted important aspects for the user and the input, the LLM selects the optimal response from the set of candidate responses generated through multi-directional thinking. This is guided by the prompt shown in Figure~\ref{fig:best-of-N-prompt}.
    
    \item \textit{Mixture-of-N:} Given the extracted important aspects for the user and the input, the LLM integrates different aspects of the candidate responses from the set of multi-directional thinking outputs, guided by the prompt in Figure~\ref{fig:mixture-of-N-prompt}. This method emulates human problem-solving by exploring multiple solutions and synthesizing a new response that leverages the strengths of all candidates.
\end{itemize}

\begin{figure}
    \centering
    \includegraphics[width=0.9\linewidth]{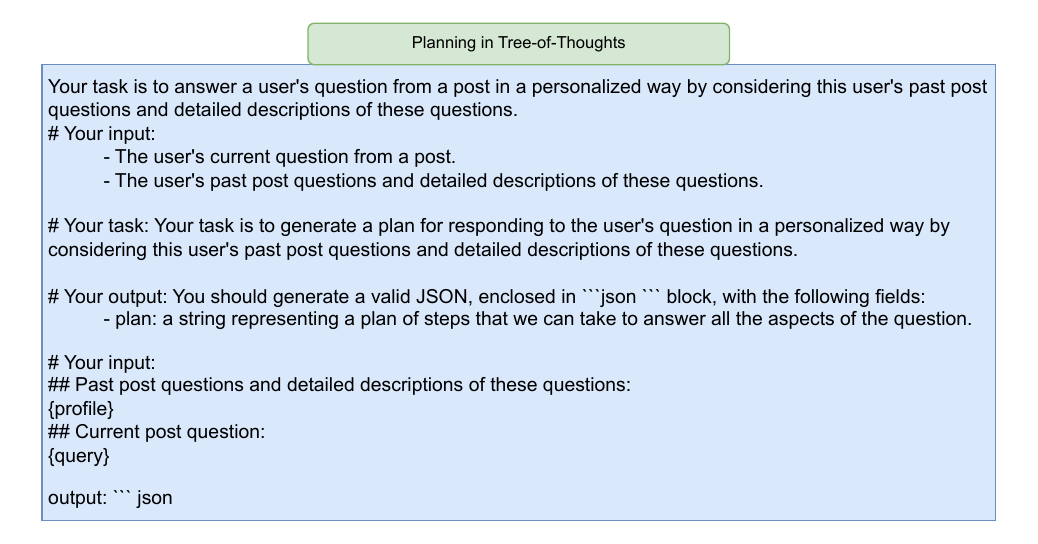}
    \vspace{-0.6cm}
    \caption{Prompt for the planning step in ToT.}
    \label{fig:plan-tot}
\end{figure}

\section{Baselines Prompts}
\label{app:baselines}

\begin{figure}
    \centering
    \includegraphics[width=0.9\linewidth]{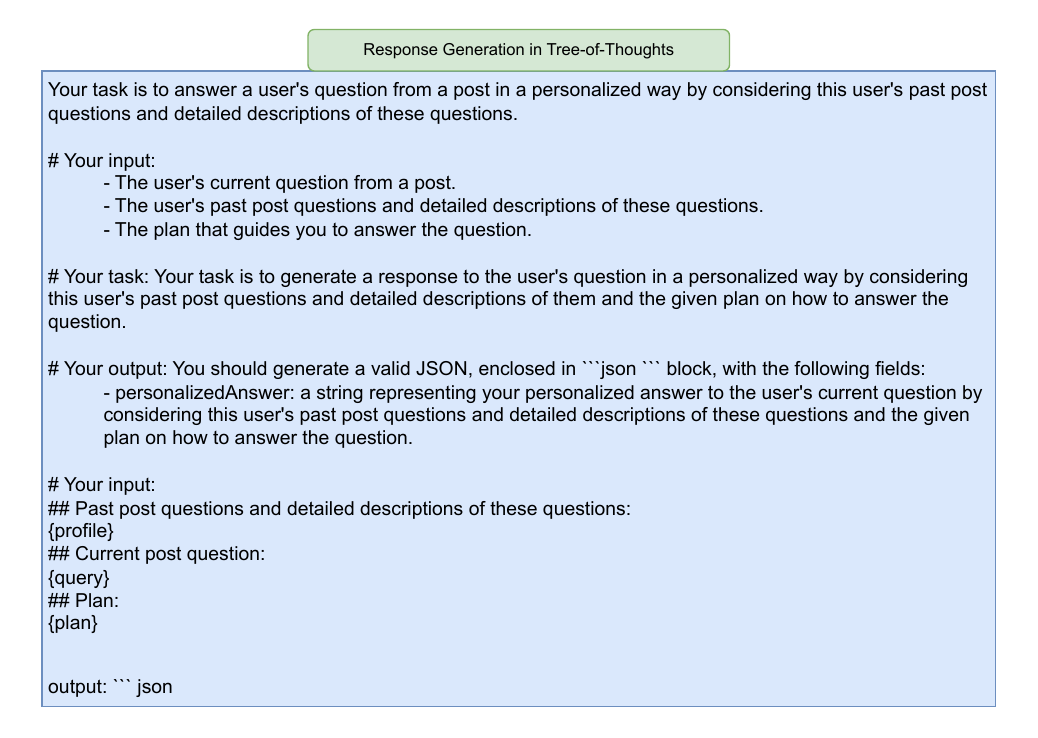}
    \vspace{-0.6cm}
    \caption{Prompt for the response generation step in ToT.}
    \label{fig:gen-tot}
\end{figure}

\begin{figure}
    \centering
    \includegraphics[width=0.9\linewidth]{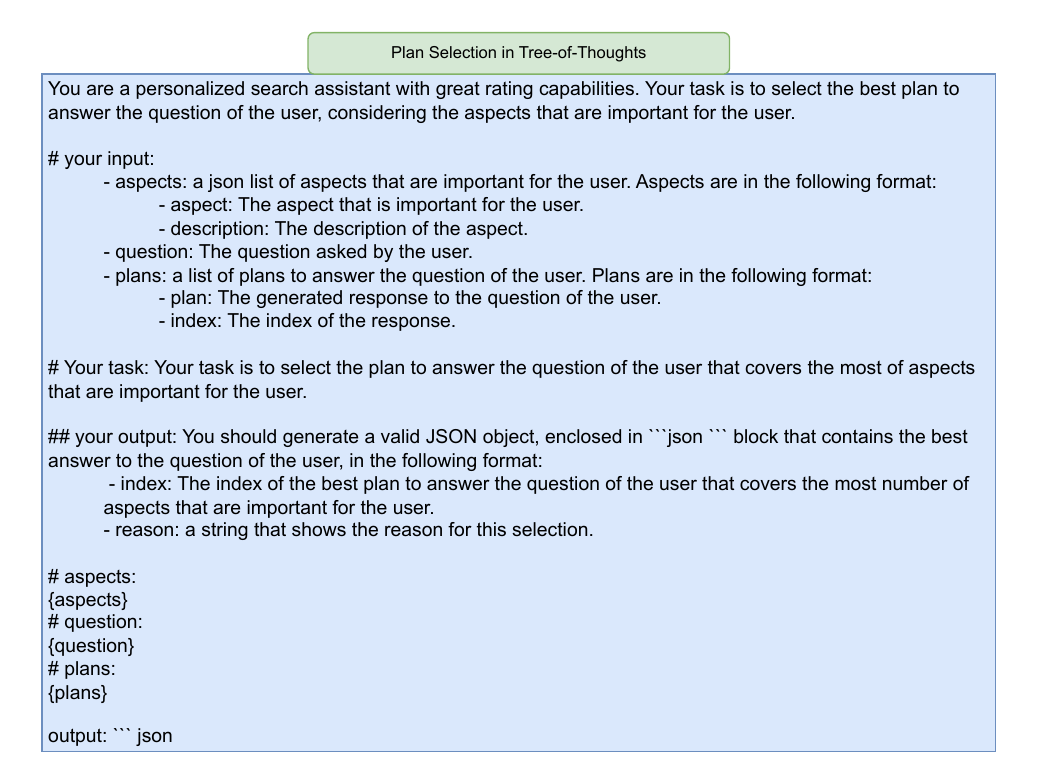}
    \vspace{-0.6cm}
    \caption{Prompt for the plan selection step in Tree of Thoughts \cite{tot}. The important aspects extracted for the user are generated using the prompt shown in Figure~\ref{fig:important-aspects-user-prompt}.}
    \label{fig:select-plan-tot}
\end{figure}

Figures~\ref{fig:baseline} and \ref{fig:baseline-cot} present the prompts used for the personalized baselines without and with chain-of-thought reasoning, respectively. For the Tree of Thoughts baseline, Figure~\ref{fig:plan-tot} shows the planning prompt, Figure~\ref{fig:gen-tot} illustrates the response generation prompt, and Figure~\ref{fig:select-plan-tot} provides the plan selection prompt.

\begin{figure*}
    \centering
    \includegraphics[width=0.96\textwidth]{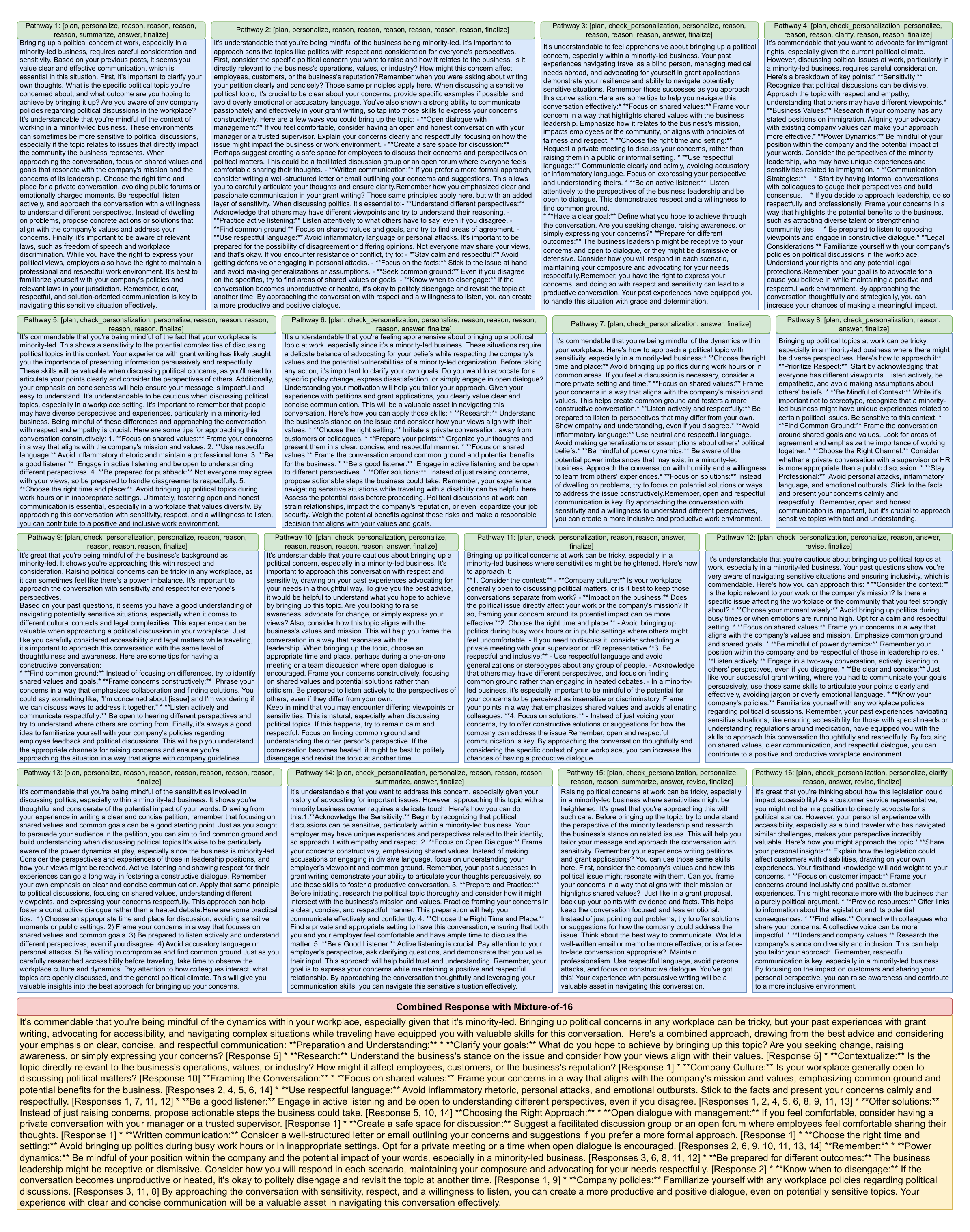}
    \vspace{-0.8cm}
    \caption{A case study of \rmdpwithsearch with $N=16$ and $T=8$ for question: What's the best way for me to bring up a topic of political concern if the business I work for is minority-led?}
    \label{fig:case-study}
\end{figure*}

\end{document}